\documentclass[oneside,11pt]{article}

\usepackage{times,url,mathrsfs}
\usepackage{amsmath,wrapfig,color}
\usepackage{amsfonts}
\usepackage{graphicx}
\usepackage{subfigure}
\newtheorem{theorem}{Theorem}
\newtheorem{lemma}{Lemma}

\begin{document}

\title{Sign Stable Projections, Sign Cauchy Projections and Chi-Square Kernels}
\author{ \bf{Ping Li} \\
         Department of Statistics \& Biostatistics\\
         Department of Computer Science\\
       Rutgers University\\
          Piscataway, NJ 08854\\
       \texttt{pl314@rci.rutgers.edu}\\\\
       \and
         \bf{Gennady Samorodnitsky}\\
         School of ORIE and Department of Statistical Science\\
         Cornell University\\
         Ithaca, NY 14853\\
         \texttt{gs18@cornell.edu}
       \and
         \bf{John Hopcroft}\\
         Department of Computer Science\\
         Cornell University\\
         {Ithaca, NY 14853}\\
        \texttt{jeh@cs.cornell.edu}}

\date{}

\maketitle

\begin{abstract}

The method of {\em stable random projections}  is popular for efficiently computing the $l_\alpha$ distances  in high dimension (where $0<\alpha\leq2$), using small space. Because it adopts nonadaptive linear projections, this method is naturally suitable when the data are collected in a dynamic streaming fashion (i.e., {\em turnstile} data streams). In this paper, we propose to use only the signs of the projected data and analyze the probability of collision (i.e., when the two signs differ). We derive a bound of the collision probability which is exact when $\alpha=2$ and becomes less sharp when $\alpha$ moves away from 2. Interestingly, when $\alpha=1$ (i.e., Cauchy random projections), we  show that the  probability of collision can be accurately approximated as functions of the chi-square ($\chi^2$) similarity. For example, when the (un-normalized) data are binary, the maximum approximation error of the collision probability is smaller than $0.0192$.  In text and vision applications, the $\chi^2$ similarity is a popular  measure for nonnegative data when the features are generated from histograms. Our experiments  confirm that the proposed method is promising for large-scale  learning applications.

\end{abstract}

\newpage

\section{Introduction}

High-dimensional representations have become very popular in modern applications of machine learning, computer vision, and information retrieval. Given two high-dimensional data vectors (e.g., two images or Web pages) $u, v\in \mathbb{R}^D$, a  basic task  is to compute their {\em distance} or {\em similarity}. For example, the correlation ($\rho_2$) and $l_2$ distance ($d_2$) are commonly used:
\begin{align}
&\rho_2(u,v) = \frac{\sum_{i=1}^D u_iv_i}{\sqrt{\sum_{i=1}^Du_i^2\sum_{i=1}^Dv_i^2}},\hspace{0.5in} d_2(u,v) = \sum_{i=1}^D |u_i-v_i|^2
\end{align}
The definition of $d_2$  naturally extends to the $l_\alpha$ distance $d_\alpha$:
\begin{align}
d_\alpha(u,v) = \sum_{i=1}^D |u_i-v_i|^\alpha
\end{align}
Here, $\alpha$ can often be viewed as {\em tuning} parameter in practice. It is known that when $0<\alpha\leq2$, the commonly used  (generalized) radial basis kernel defined as $K(u,v) = \exp\left(-\lambda d_\alpha(u,v)\right)$, where $\lambda>0$, is positive definite~\cite{Book:Scholkopf_02}.\\

In this study, we are particularly interested in the $\chi^2$ similarity, denoted by $\rho_{\chi^2}$:
\begin{align}\label{eqn_rho_chi2}
&\rho_{\chi^2} = \sum_{i=1}^D \frac{2u_iv_i}{u_i+v_i},\hspace{0.2in}\text{where } \  \ u_i\geq 0,\ v_i\geq 0, \ \ \ \sum_{i=1}^D u_i = \sum_{i=1}^D v_i=1
\end{align}
The chi-square similarity is closely related to the  chi-square distance $d_{\chi^2}$:
\begin{align}
&d_{\chi^2} = \sum_{i=1}^D \frac{(u_i-v_i)^2}{u_i+v_i}  = \sum_{i=1}^D (u_i+v_i) -\sum_{i=1}^D \frac{4u_iv_i}{u_i+v_i} = 2-2\rho_{\chi^2}
\end{align}
The  chi-square similarity is an instance of   the Hilbertian metrics, which are defined over  probability space~\cite{Proc:Hein_AISTATS05} and hence suitable for data generated from histograms. Histogram-based features (e.g., bag-of-word or bag-of-visual-word models) are extremely popular in computer vision, natural language processing (NLP), and information retrieval. Empirical studies have demonstrated the superiority of the $\chi^2$ distance over  $l_2$ or $l_1$ distances for image and text classification tasks~\cite{Article:Chapelle_99,Proc:Hein_AISTATS05,Proc:Jiang_CIVR07,Proc:Alexe_CVPR10,Proc:Wang_CVPR09,Proc:Vempati_BMVC10,Article:Vedaldi_PAMI12}.\\

The current trend  in machine learning is to use  ultra-high-dimensional representations. For example, Winner of 2009 PASCAL image classification challenge  used {4 million} features~\cite{Proc:Wang_CVPR10}. \cite{Url:Goog_Large_Learning,Proc:Weinberger_ICML2009} described applications using billion or trillion features. The use of extremely high-dimensional data often produces good accuracies but the cost is the significant increase in computations, storage, and energy consumptions.

The method of {\em normal random projections} (i.e., $\alpha$-stable projections with $\alpha=2$) has become popular in machine learning (e.g.,~\cite{Proc:Frund_NIPS08,Proc:Li_Hastie_Church_COLT06}) for reducing the data dimensions and data sizes, to facilitate efficient computations of  the $l_2$ distances and correlations.  More generally, the method of {\em stable random projections}~\cite{Article:Indyk_JACM06,Proc:Li_SODA08} provides an efficient algorithm to compute the $l_\alpha$ distances, for $0<\alpha\leq2$. \\

In this paper, we propose to use only the signs of the projected data after applying $\alpha$-stable projections on the original data. In particular, we show that when $\alpha=1$, the probability of collision (i.e., when the two signs differ) is closely related to the $\chi^2$ similarity. Thus, our method provides an effective strategy for large-scale machine learning when the applications favor the use of $\chi^2$ similarity.

\subsection{Stable Random Projections and  Sign (1-Bit) Stable Random Projections }

Consider two high-dimensional vectors $u, v\in\mathbb{R}^D$.  The basic  idea of stable random projections is to multiply $u$ and $v$ by a random matrix $\mathbf{R}\in\mathbb{R}^{D\times k}$: $x = u\mathbf{R}\in\mathbb{R}^k$, $y = v\mathbf{R}\in\mathbb{R}^k$, where entries of $\mathbf{R}$ are i.i.d. samples from a symmetric $\alpha$-stable distribution with unit scale. By properties of stable distributions,  $x_j-y_j$  follows a symmetric $\alpha$-stable distribution with scale   $d_\alpha$.  Hence, the task of computing $d_\alpha$ boils down to estimating the scale  $d_\alpha$ from $k$ i.i.d. samples.

In this paper, we propose to store only the signs of projected data and study the  probability of collision:
\begin{align}
P_\alpha=\mathbf{Pr}\left(\text{sign}(x_j)\neq \text{sign}(y_j)\right)
\end{align}
Using only the signs (i.e., 1 bit) has significant advantages for  applications in search and learning.
When $\alpha=2$, this probability can be analytically evaluated~\cite{Article:Goemans} (or via a simple geometric argument):
\begin{align}
P_2 = \mathbf{Pr}\left(\text{sign}(x_j)\neq \text{sign}(y_j)\right) = \frac{1}{\pi}\cos^{-1}\rho_2
\end{align}
which is an important result known as {\em sim-hash}~\cite{Proc:Charikar}. For $\alpha<2$, the collision probability is an open problem. When the data are nonnegative, this paper will prove a bound  of $P_\alpha$ for general $0<\alpha\leq2$:
\begin{align}\label{eqn_rho_alpha}
P_\alpha=\mathbf{Pr}\left(\text{sign}(x) \neq \text{sign}(y)\right)
\leq &\frac{1}{\pi}\cos^{-1}\rho_\alpha, \hspace{0.1in} \text{where }  \ \ \rho_\alpha = \left(\frac{\sum_{i=1}^D u_i^{\alpha/2}v_i^{\alpha/2}}{\sqrt{\sum_{i=1}^Du_i^\alpha\sum_{i=1}^D v_i^\alpha}}\right)^{2/\alpha}
\end{align}
This bound is exact at $\alpha=2$ and  less sharp as $\alpha$ moves away from 2. Furthermore, for $\alpha=1$ and nonnegative data, we have the interesting observation  that the  probability $P_1$ can be well approximated as functions  of the $\chi^2$ similarity $\rho_{\chi^2}$.

\subsection{The Advantages of Sign Stable Random Projections}

The advantage of {\em sign stable random projections} can be summarized as follows:
\begin{enumerate}
\item There is a significant saving in storage space by using only 1 bit instead of (e.g.,) 64 bits.

\item This scheme leads to an efficient linear algorithm (e.g., linear SVM). For example, a negative sign can be coded as ``01'' and a positive sign as ``10'' (i.e., a vector of length 2). With $k$ projections, we  concatenate   $k$ short vectors to form a vector of length $2k$. This idea is inspired by  {\em b-bit minwise hashing}~\cite{Proc:Li_Konig_WWW10,Proc:Li_Owen_Zhang_NIPS12}, which is designed for binary data and the ``resemblance'' similarity.

\item This scheme also leads to an efficient near neighbor search algorithm~\cite{Article:Friedman_75}. For this application, we can code a negative sign by ``0'' and positive sign by ``1'' and concatenate $k$ such bits to form a hash table of $2^k$ buckets. In the query phase, one only  searches for similar vectors in one bucket. Usually $L$ such tables are independently constructed  to increase the accuracy.
    This provides  a simple and efficient implementation of the general framework of locality sensitive hashing~\cite{Proc:Indyk_STOC98}. For this paper, we will not focus on near neighbor search.

\end{enumerate}

\subsection{Data Stream Computations}

Stable random projections are naturally suitable for {\em data stream} computations. In modern applications,  massive datasets are  often generated in a streaming fashion, which  are difficult to transmit and store~\cite{Article:Muthukrishnan_05}, as the processing
is  done \textbf{on the fly} in \textbf{one-pass} of the data. The problem of ``scaling up for high dimensional data and high speed data streams'' is among the  ``ten challenging problems in data mining research''~\cite{Article:ICDM10}. Network traffic is a typical example of data streams and network data can easily reach petabyte scale~\cite{Article:ICDM10}.

In the standard {\em turnstile} model~\cite{Article:Muthukrishnan_05}, a data stream can be viewed as high-dimensional vector with the entry values changing over time. Here, we denote a stream at time $t$ by  $u_i^{(t)}$, $i = 1$ to $D$. At time $t$, a stream element $(i_{t}, I_t)$ arrives and updates the $i_{t}$-th coordinate as
\begin{align}\label{eqn_Turnstile}
u_{i_{t}}^{(t)} =  u_{i_t}^{(t-1)} + I_{t}
\end{align}
Clearly, the turnstile data stream model is particularly suitable for describing histograms and it is also a standard model for network traffic summarization and monitoring~\cite{Proc:Zhao_NPC11}.  Because this stream model is linear, methods based on linear projections (i.e., matrix-vector multiplications) can naturally handle streaming data of this sort. Basically, entries of the projection matrix $\mathbf{R}\in\mathbb{R}^{D\times k}$  are (re)generated as needed using pseudo-random number techniques~\cite{Proc:Nisan_STOC90}. At the element $(i_t, I_t)$ arrives, only the entries in the $i_t$-th row, i.e., $r_{i_t,j}$, $j=1$ to $k$, are (re)generated and the projected data  are updated as
\begin{align}
x_j^{(t)} = x_j^{(t-1)} + I_t \times r_{i_t j}, \ \ \ j = 1, 2, ..., D
\end{align}
Thus,  the original data, e.g., $u, v\in\mathbb{R}^D$, are not stored, as a significant advantage of using linear  projections. \\

Recall that, in the definition of $\chi^2$ similarity, the data are assumed to be normalized (summing to 1). For nonnegative data streams, the sum can be computed error-free by using merely one counter: $\sum_{i=1}^D u_i^{(t)} = \sum_{s=1}^t I_s$. This means that we can still use, without loss of generality, the sum-to-one assumption, even in the streaming environment. This fact was recently exploited  by another data stream algorithm named {\em Compressed Counting (CC)}~\cite{Proc:Li_Zhang_COLT11} for  efficiently estimating the {\em Shannon entropy} of data streams. \\

Because the $\chi^2$ similarity is very popular in computer vision, recently there are proposals for estimating the $\chi^2$ similarity. For example, \cite{Report:Fuxin2013} proposed a nice technique to approximate $\rho_{\chi^2}$ by first expanding the data from $D$ dimensions to (e.g.,) $5\sim 10 \times D$ dimensions through a {\em nonlinear} transformation and then applying normal random projections on the expanded data. Because the nonlinear operations must be applied to the original data, the method is not applicable to data stream computations. \\

For notational simplicity,  we will drop the  superscript $(t)$ for the rest of the paper.

\subsection{Paper Organization}

In Section~\ref{sec_chi2_exp}, we provide a set of SVM  experiments to illustrate the potential advantages of the use of $\chi^2$ similarity $\rho_{\chi^2}$ compared to linear kernels. For example, on the {\em MNIST-small} dataset, using $\chi^2$ kernels can achieve significantly more accurate classification results than using linear kernels, without using additional tuning parameter.

In Section~\ref{sec_bound}, we derive a general bound of the collision probability for sign stable random projections, for  all $0<\alpha\leq2$. As verified by simulations, this bound is exact when $\alpha=2$ and becomes less sharp as $\alpha$ moves away from 2.  In Section~\ref{sec_Cauchy}, we focus on Cauchy random projections (i.e., $\alpha=1$) and analyze several properties of the $\chi^2$ similarity.

In Section~\ref{sec_approximation}, we propose two approximations of the collision probability for $\alpha=1$, as functions of the $\chi^2$ similarity. We prove that when the data are binary, the approximation error would be smaller than $0.0192$. Furthermore, we validate  the proposed approximations on a dataset of  3.6 million pairs of word occurrence vectors. Section~\ref{sec_classification} provides an experimental study for SVM classifications using our proposed approximations. Finally, Section~\ref{sec_conclusion} concludes the paper and also points out that the processing cost of stable random projections can be substantially reduced by using a {\em very sparse} projection matrix.

\newpage\clearpage

\section{An Experimental Study of Chi-Square Kernels}\label{sec_chi2_exp}

We   provide an experimental study to validate the use of $\chi^2$ similarity. Here, the ``$\chi^2$-kernel'' is defined as  $K(u,v) = \rho_{\chi^2}$ and the ``acos-$\chi^2$-kernel''  as $K(u,v) = 1-\frac{1}{\pi}\cos^{-1}\rho_{\chi^2}$. With a slight abuse of terminology, we call both ``$\chi^2$ kernel'' when it is clear in the context.

 We use the ``precomputed kernel'' functionality in LIBSVM  on two datasets: (i) {\em UCI-PEMS}, with 267 training examples and 173 testing examples in 138,672 dimensions; (ii) {\em MNIST-small}, a subset of the popular {\em MNIST} dataset, with 10,000 training examples and 10,000 testing examples.\footnote{When we used the original MNIST data (60,000 training examples) using LIBSVM ``precomputed kernel'' functionality (which is memory intensive), we observed seg fault in matlab on a workstation with 96GB RAM. }\\

\begin{figure}[h!]
\begin{center}
\includegraphics[width = 2.8in]{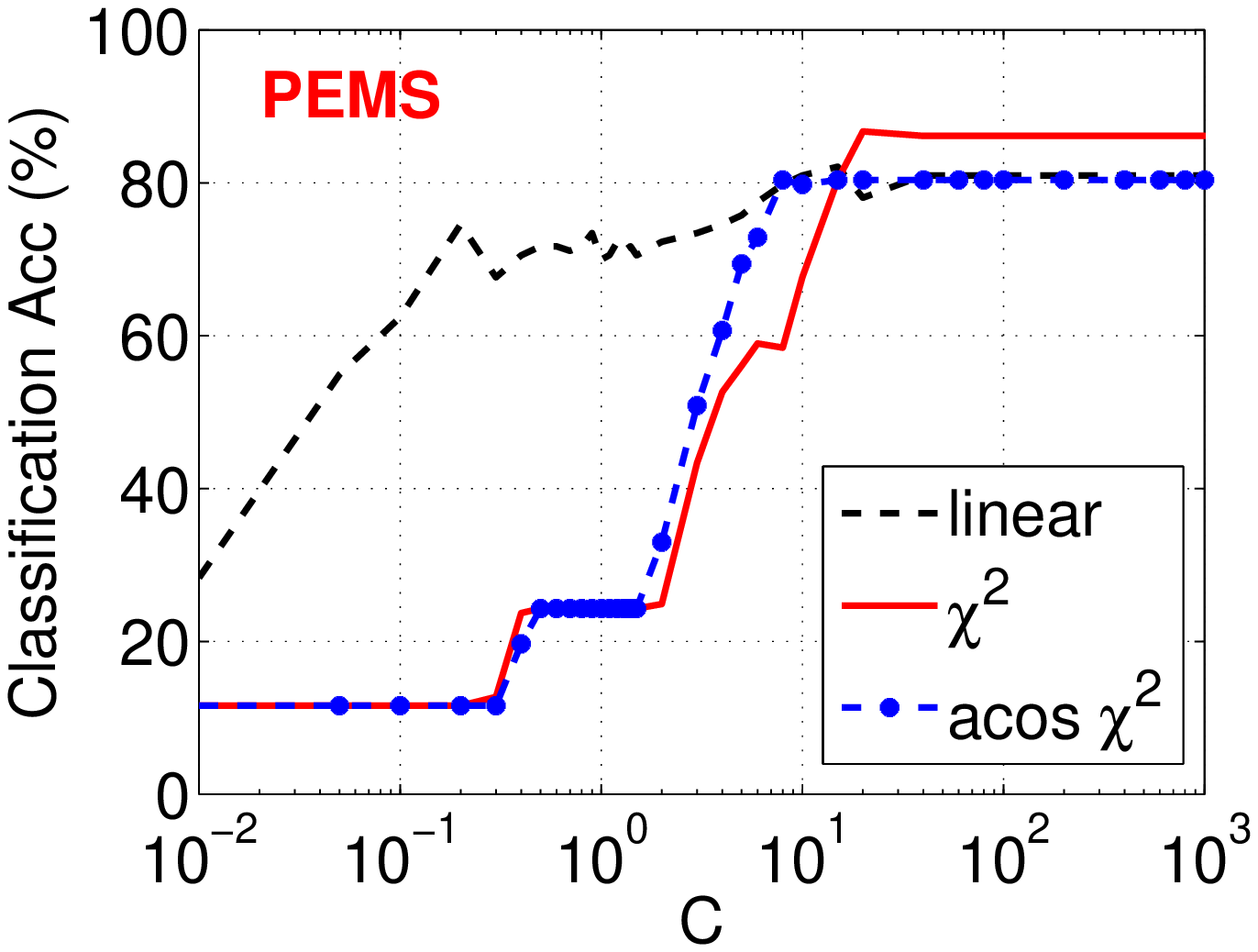}\hspace{0.3in}
\includegraphics[width = 2.8in]{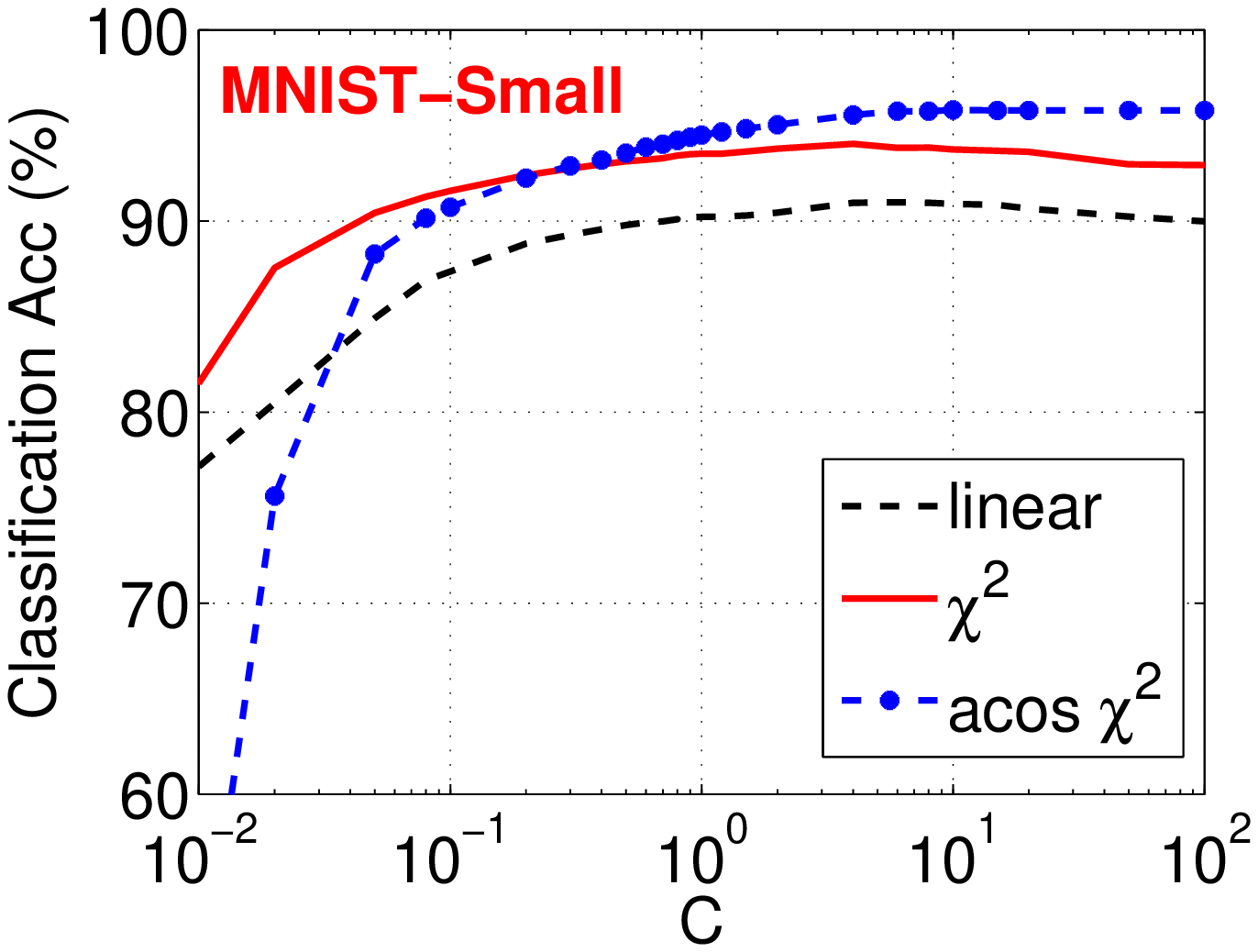}
\end{center}
\vspace{-0.2in}
\caption{Classification accuracies. $C$ is the $l_2$-regularization parameter. We use LIBLINEAR for ``linear'' (i.e., $\rho_2$) kernel and LIBSVM ``precomputed kernel'' functionality for two types of $\chi^2$ kernels (``$\chi^2$-kernel'' and ``acos-$\chi^2$-kernel''). For {\em UCI-PEMS},  the $\chi^2$-kernel has better performance than the linear kernel and acos-$\chi^2$-kernel. For {\em MNIST-Small}, both $\chi^2$ kernels noticeably outperform linear kernel. Note that, although we place three curves together, we do not mean one should compare them at a fixed $C$. We assume the best $C$ could be found by (e.g.,) cross-validations for each method. The {\em MNIST-small} dataset used the original testing examples and only 1/6 of the original training examples.  }\label{fig_base_chi2}
\end{figure}

The results are shown in Figure~\ref{fig_base_chi2}. To compare these two types of $\chi^2$ kernels with ``linear'' kernel, we also test the same data using LIBLINEAR~\cite{Article:Fan_JMLR08} after normalizing the data to have unit Euclidian norm, i.e., we basically use  $\rho_2$. For both LIBSVM and LIBLINEAR, we use $l_2$-regularization with a parameter $C$ and we report the test errors for a wide range of $C$ values.

Here, we should state that,  while the use of $\chi^2$ similarity has become popular in computer vision, it is not the intention of this paper to use these two small examples to conclude the advantage of $\chi^2$ kernels over linear kernel. We  simply use them to validate our method for approximating the $\chi^2$ similarity, as our proposed method is general-purpose, not limited to data generated from histograms.\\

The {\em MNIST-small} dataset is  more close to computer vision applications. Using linear kernels, the classification accuracy is about $90\%$ at its best $C$ value, while using the $\chi^2$ kernels can achieve about $95\%$. Note that our proposed method will be able to approximate the acos-$\chi^2$-kernel by linear kernels. In other words, our method may be able to provide both the computational efficiency of linear kernels and good accuracy of $\chi^2$ kernels (in certain datasets). Unlike the RBF nonlinear kernel, the $\chi^2$ kernels do not use additional tuning parameters (other than $C$). Since {\em MNIST-small} used only $1/6$ of the original training examples and the same testing examples, the performance of $\chi^2$ kernels on this dataset is impressive.


\section{Sign Stable Random Projections and the Collision Probability Bound}\label{sec_bound}

Consider two high-dimensional vectors $u, v\in\mathbb{R}^D$. We apply stable random projections:
\begin{align}
x = \sum_{i=1}^D u_i r_i,\ \  \ y = \sum_{i=1}^D v_i  r_i,\ \   \ r_i \sim S(\alpha,0,1), i.i.d.
\end{align}
Here $Z\sim S(\alpha,\beta,\gamma)$ denotes an $\alpha$-stable $\beta$-skewed distribution with scale $\gamma>0$. When $\beta=0$ (i.e., symmetric), its characteristic function~\cite{Book:Samorodnitsky_94} is
$E\left(e^{\sqrt{-1}Zt}\right) = e^{-\gamma |t|^\alpha}$. By properties of stable distributions, we know
$x-y \sim S\left(\alpha,0,\sum_{i=1}^D|u_i-v_i|^\alpha\right)$.  This method transforms the task of computing the $l_\alpha$ distance $\sum_{i=1}^D |u_i - v_i|^\alpha$ into a problem of estimating the scale parameter from stable samples. Note that in this section, we only use a projection vector $\mathbf{R}\in\mathbb{R}^{D\times k}$ with $k=1$, because the samples are i.i.d. anyway. Consequently, we simply use $x$ and $y$ to denote the projected instead of $x_j$ and $y_j$.\\

Applications including linear learning and near neighbor search will benefit from {\em sign $\alpha$-stable random projections}. When $\alpha=2$, the
{ collision probability} $\mathbf{Pr}\left(\text{sign}(x) \neq \text{sign}(y)\right)$ is known~\cite{Proc:Charikar,Article:Goemans}. For $\alpha<2$,  it is a difficult probability problem. In this section, we will study a bound  of $\mathbf{Pr}\left(\text{sign}(x) \neq \text{sign}(y)\right)$ for general $0<\alpha\leq2$, which is fairly accurate for $\alpha$ close to 2.

\subsection{Collision Probability Bound}

In this paper, we focus on nonnegative data (as common in practice). We present our first theorem.
\begin{theorem}\label{thm_upper_bound}
When the data are nonnegative,  $u_i\geq0, v_i\geq 0$, we have
\begin{align}
\mathbf{Pr}\left(\text{sign}(x) \neq \text{sign}(y)\right) =& E\left(\frac{1}{\pi}\cos^{-1}\left(\frac{\sum_{i=1}^D u_iv_iT_i}{\sqrt{\sum_{i=1}^Du_i^2T_i\sum_{i=1}^D v_i^2T_i}}\right)\right)\\
\leq &\frac{1}{\pi}\cos^{-1}\left\{\left(\frac{\sum_{i=1}^D u_i^{\alpha/2}v_i^{\alpha/2}}{\sqrt{\sum_{i=1}^Du_i^\alpha\sum_{i=1}^D v_i^\alpha}}\right)^{2/\alpha}\right\}
=\frac{1}{\pi}\cos^{-1}\rho_\alpha
\end{align}
where $T_i \sim S\left(\alpha/2,1,\cos\left(\frac{\pi\alpha}{4}\right)\right)$, i.i.d.\\

\noindent\textbf{Proof:}\ \ See Appendix~\ref{app_thm_upper_bound}. Recall $\rho_\alpha$ is also defined in (\ref{eqn_rho_alpha}).\hfill$\Box$\\
\end{theorem}

The proof of Theorem~\ref{thm_upper_bound} utilizes an interesting  property: a symmetric stable random variable can be factorized into a product of a normal  and a maximally-skewed stable random variable~\cite{Book:Samorodnitsky_94}.
 \begin{lemma}\label{lem_factorization}
Let $X\sim S\left(\alpha^\prime,0,\gamma\right)$ with $0<\alpha^\prime\leq 2$,  $0<\alpha<\alpha^\prime$, and  $T\sim S\left(\alpha/\alpha^\prime,1,\cos\left(\frac{\pi\alpha}{2\alpha^\prime}\right)\right)$. $T$ and $X$ are independent. Then
\begin{align}
Z = T^{1/\alpha^\prime} X \sim S\left(\alpha,0,\gamma\right)
\end{align}
In particular, when $\alpha^\prime = 2$, we have
\begin{align}
Z = T^{1/2}X \sim S\left(\alpha,0,\gamma\right),\ \ \text{  where } \  \ T \sim S\left(\alpha/2,1,\cos\left(\frac{\pi\alpha}{4}\right)\right), \ \ X \sim N\left(0,2\gamma\right)
\end{align}\hfill$\Box$\\
\end{lemma}

How tight is this bound: $\mathbf{Pr}\left(\text{sign}(x) \neq \text{sign}(y)\right)\leq\frac{1}{\pi}\cos^{-1}\rho_\alpha$? The answer depends on $\alpha$. For $\alpha=2$, this bound is exact~\cite{Proc:Charikar,Article:Goemans}. In fact the result for $\alpha=2$ immediately leads to the following Lemma:
 \begin{lemma}\label{lem_rho2}
The kernel defined as $K(u,v) = 1-\frac{1}{\pi}\cos^{-1}\rho_2$ is positive definite (PD).\\

\noindent\textbf{Proof:}\ \ The indicator function $1\left\{\text{sign}(x) = \text{sign}(y)\right\}$ can be written as an inner product (hence PD) and  $\mathbf{Pr}\left(\text{sign}(x) = \text{sign}(y)\right) = E\left(1\left\{\text{sign}(x) = \text{sign}(y)\right\}\right) = 1-\frac{1}{\pi}\cos^{-1}\rho_2$.\hfill$\Box$
\end{lemma}

\subsection{A Simulation Study to Verify the  Bound of the Collision Probability}

 We  generate the original data $u$ and $v$ by sampling from a bivariate t-distribution, which has two parameters: the correlation and the number of degrees of freedom. We use a full range of the correlation parameter from 0 to 1 (spaced at 0.01). To generate positive data, we simply take the absolute values of the generated data.  Then we fix the data as our original data (like $u$ and $v$), apply sign stable random projections, and report the empirical collision probabilities (after $10^5$ repetitions). An important parameter for t-distribution is the number of degrees of freedom ($df$), which determines how heavy the tail is. For example, a t-distribution has finite expectation when $df>1$ and has finite third moment when $df>3$. We only report simulations with $df=1$ (i.e., heavy-tailed data). The results for larger $df$ values are  similar in our experiments.\\

\begin{figure}[h!]
\begin{center}
\mbox{
\includegraphics[width=2.2in]{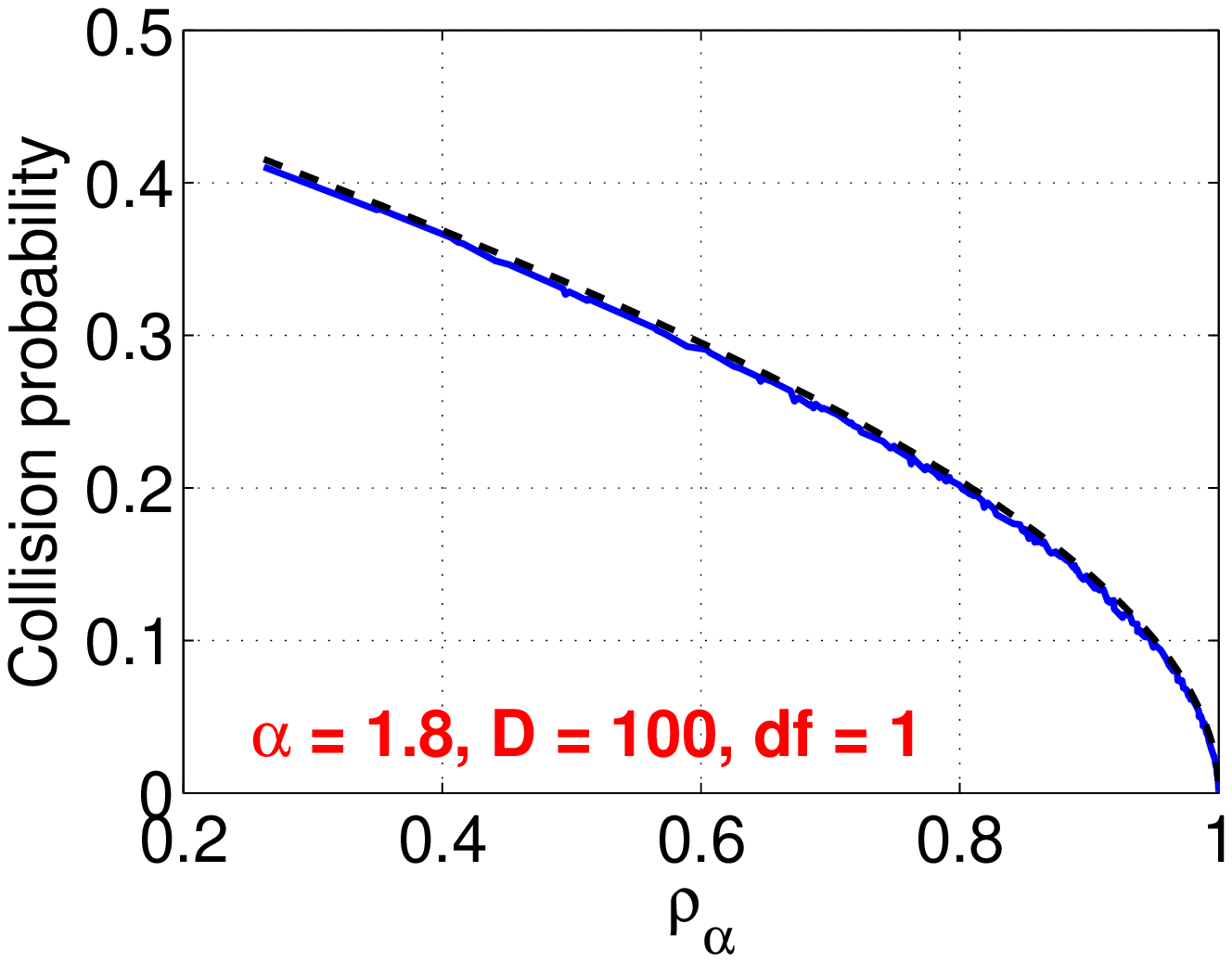}
\includegraphics[width=2.2in]{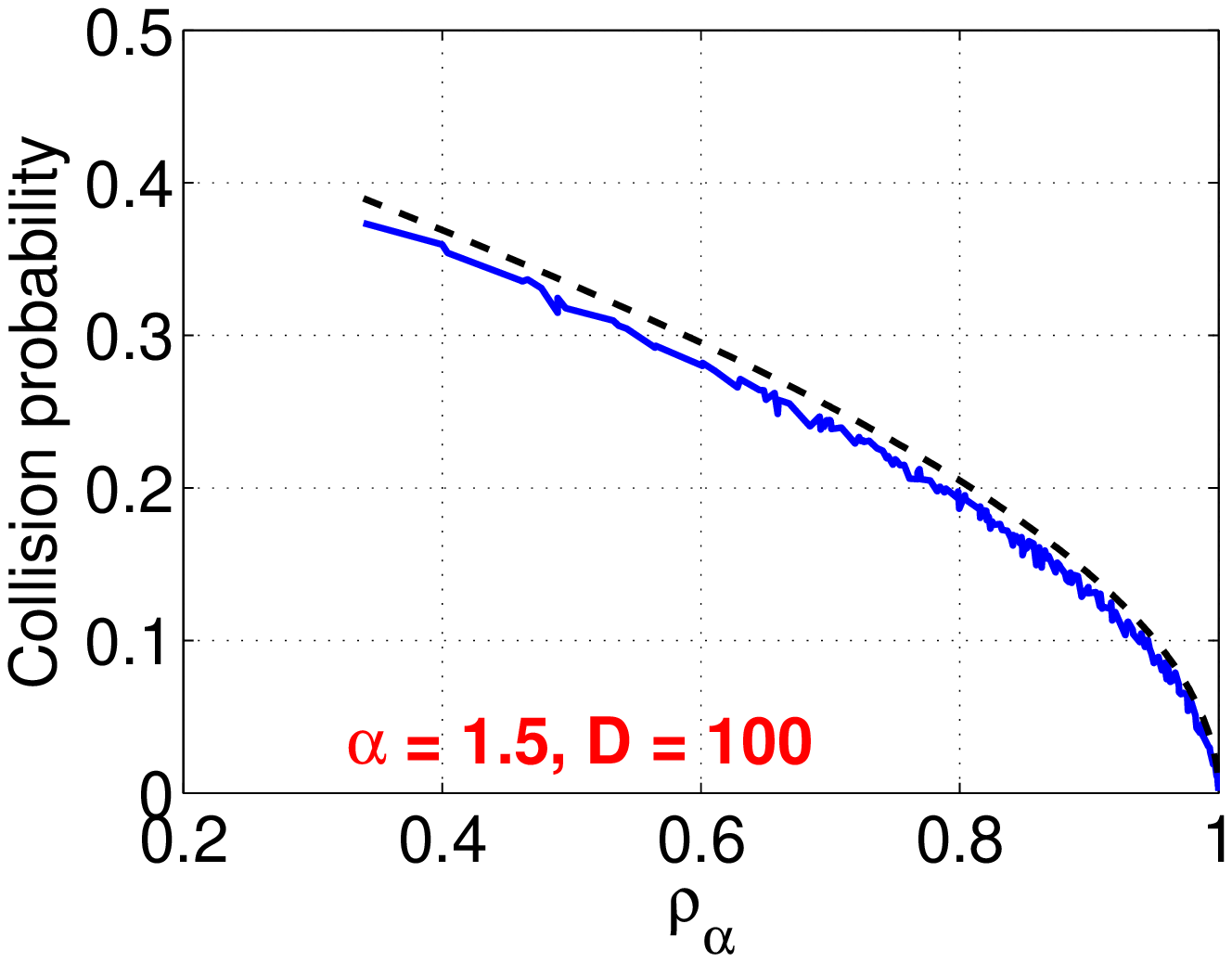}
\includegraphics[width=2.2in]{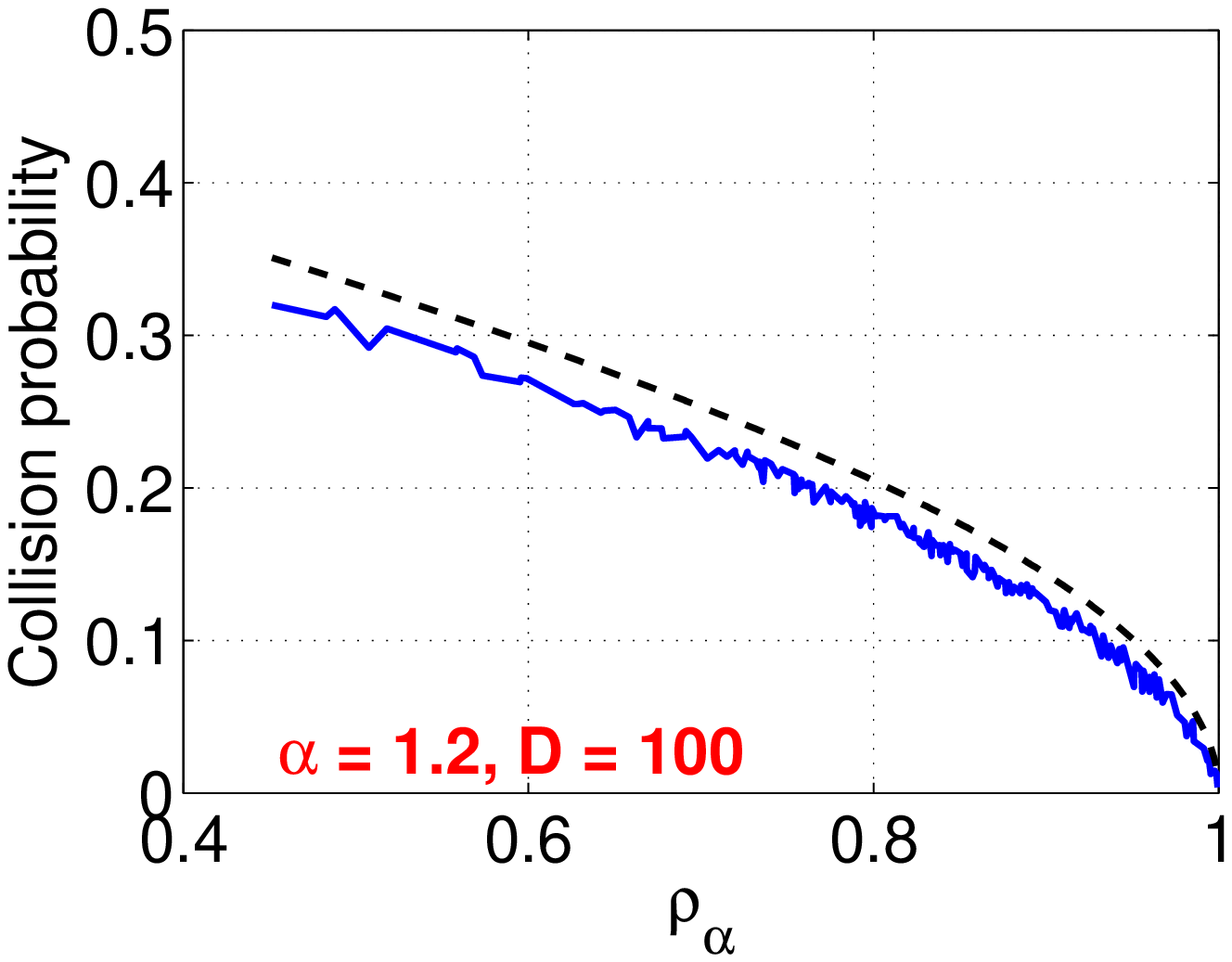}
}
\mbox{
\includegraphics[width=2.2in]{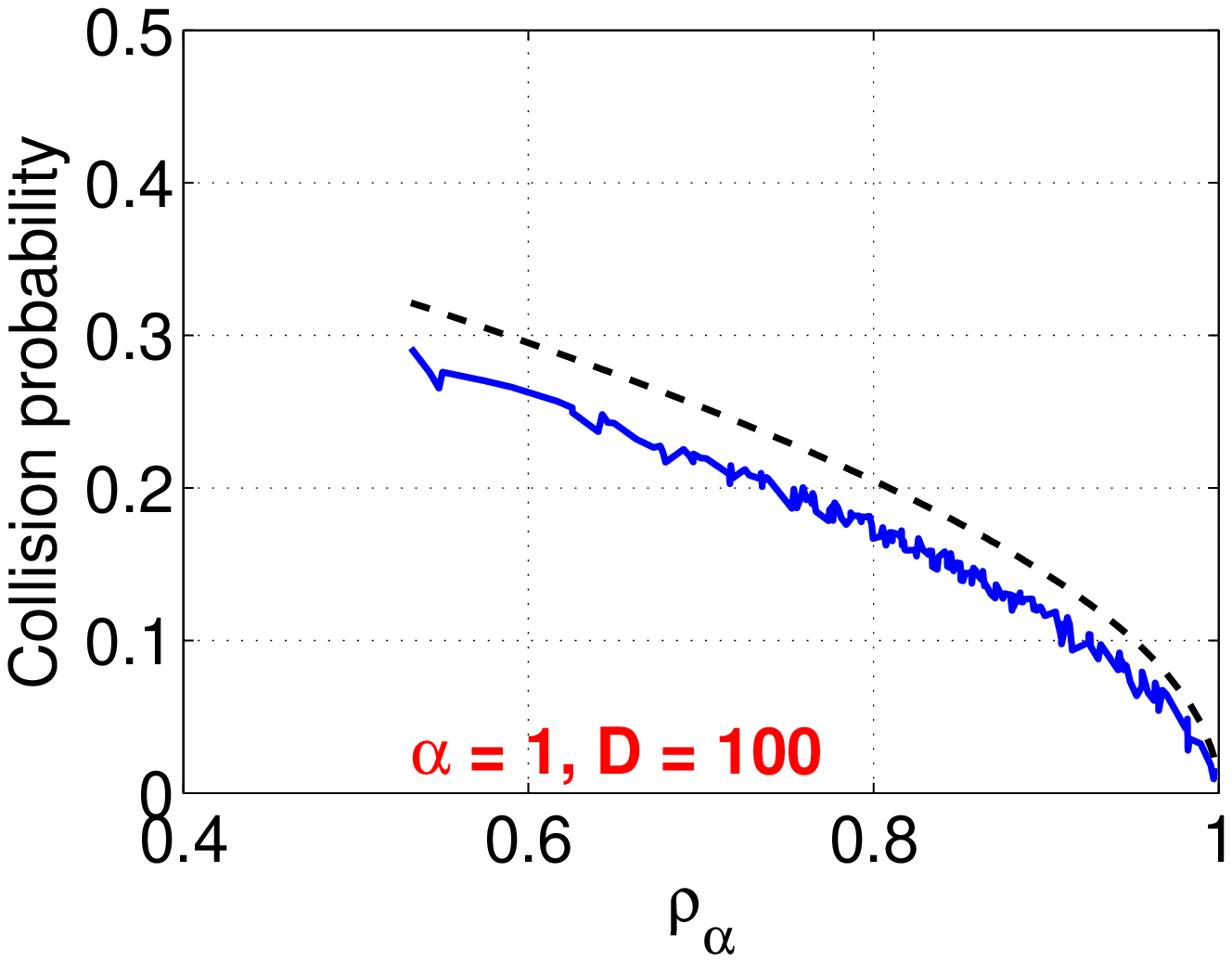}
\includegraphics[width=2.2in]{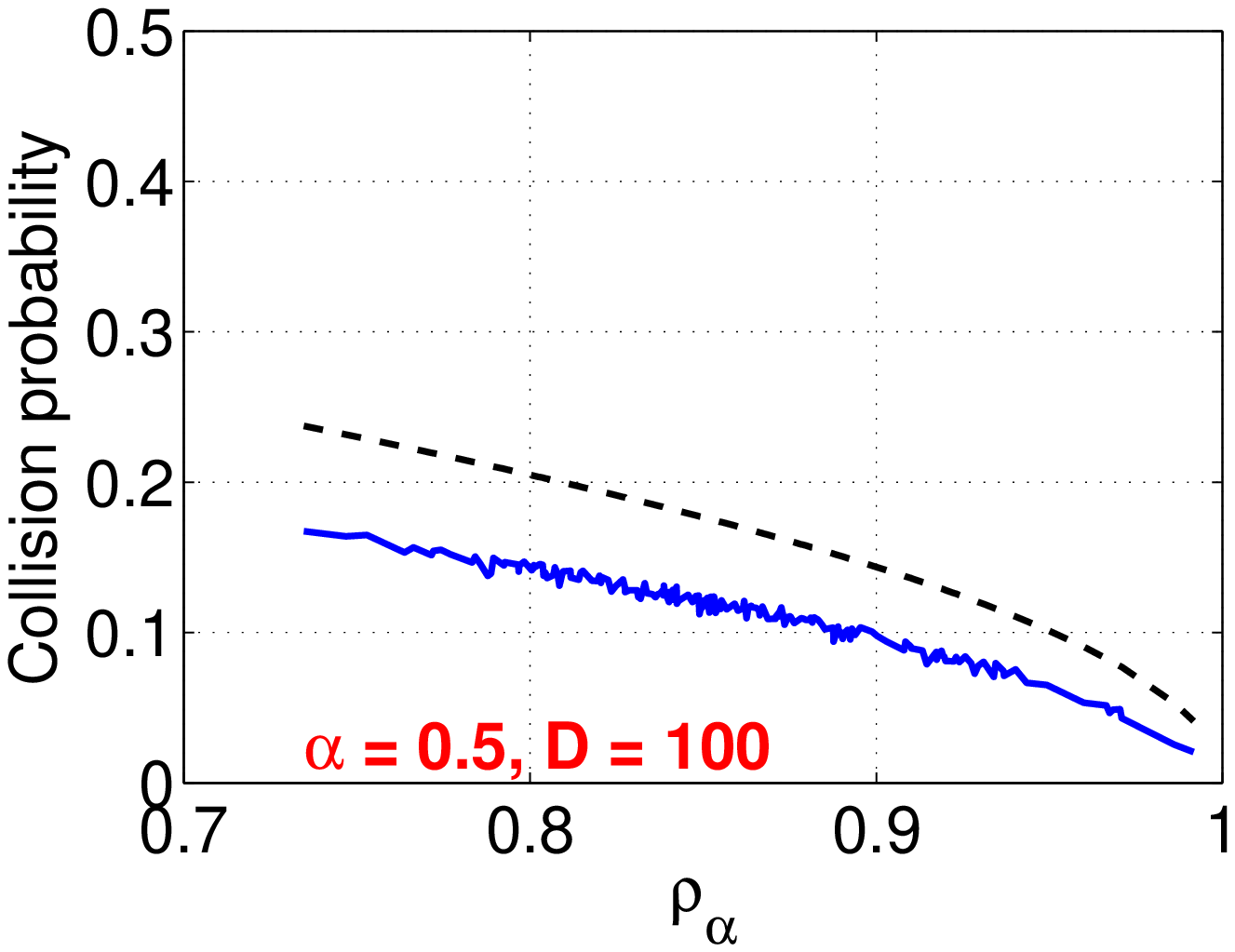}
\includegraphics[width=2.2in]{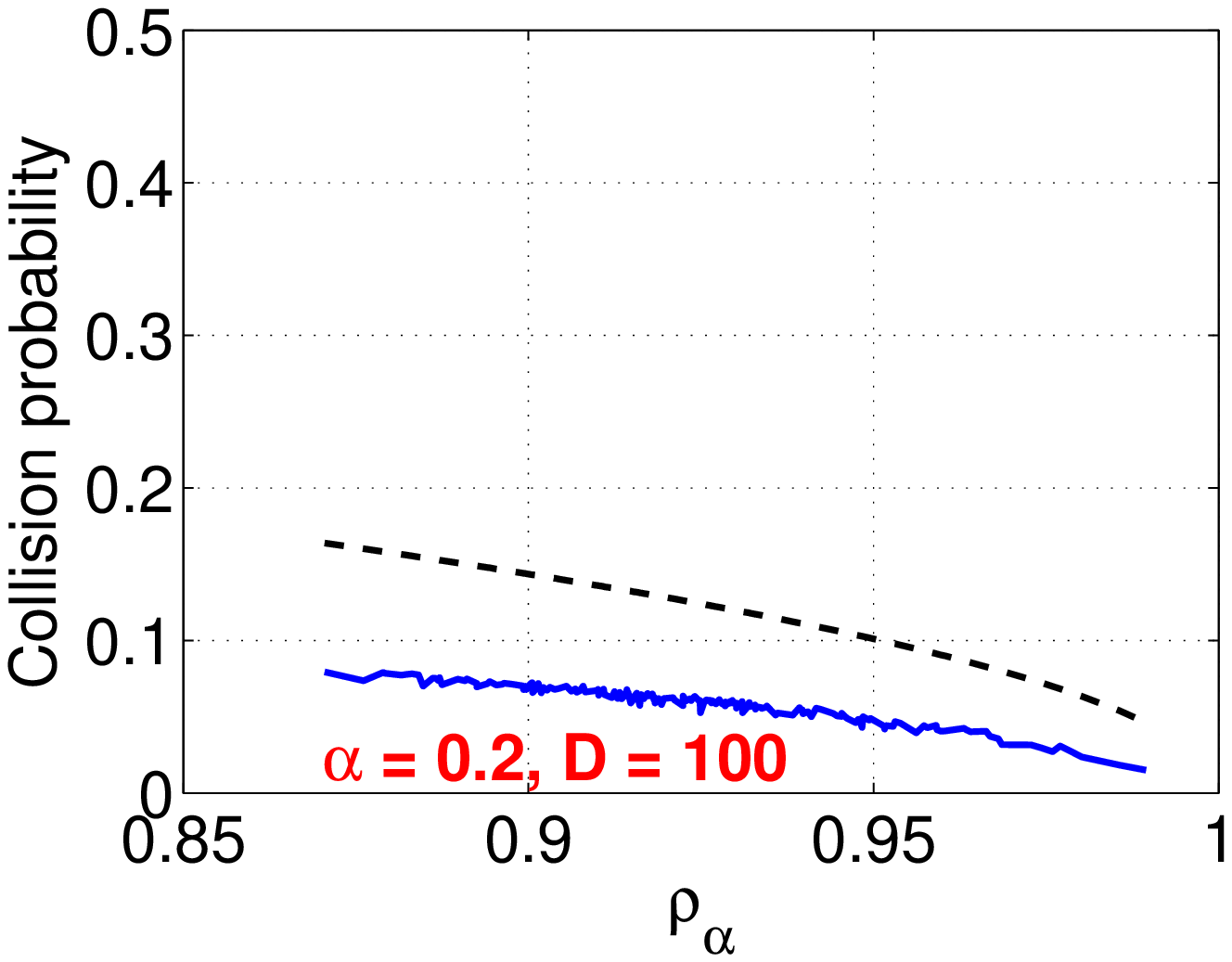}
}
\end{center}
\vspace{-0.2in}
\caption{\textbf{Dense Data} and $D=100$. Simulated collision probability $\mathbf{Pr}\left(\text{sign}(x) \neq \text{sign}(y)\right)$ for sign stable random projections. In each panel, the dashed curve is the upper bound $\frac{1}{\pi}\cos^{-1}\rho_\alpha$. }\label{fig_RhoD100T1}
\end{figure}

Figure~\ref{fig_RhoD100T1} presents  the simulated  collision probability $\mathbf{Pr}\left(\text{sign}(x) \neq \text{sign}(y)\right)$ for $D=100$ and  $\alpha\in\{1.8, 1.5, 1.2, 1.0, 0.5, 0.2\}$. In each panel, the dashed curve is the theoretical upper bound $\frac{1}{\pi}\cos^{-1}\rho_\alpha$, and the solid curve is the simulated collision probability. Note that it is expected that the simulated data can not cover the entire range of $\rho_\alpha$ values. For example, if the data are dense (no zero entries), then $\rho_\alpha\rightarrow 1$ as $\alpha\rightarrow0$. This explains why the last panel ($\alpha=0.2$) only  contains data with $\rho_\alpha>0.85$.

Figure~\ref{fig_RhoD100T1} verifies  the theoretical upper bound $\frac{1}{\pi}\cos^{-1}\rho_\alpha$. When $\alpha\geq1.5$, this upper bound is  fairly sharp. However, when $\alpha\leq 1$, the bound is not  tight, especially for small $\alpha$. Also, the curves of the empirical collision probabilities are not  smooth (in terms of $\rho_\alpha$).

\begin{figure}[h!]
\begin{center}
\mbox{
\includegraphics[width=2.2in]{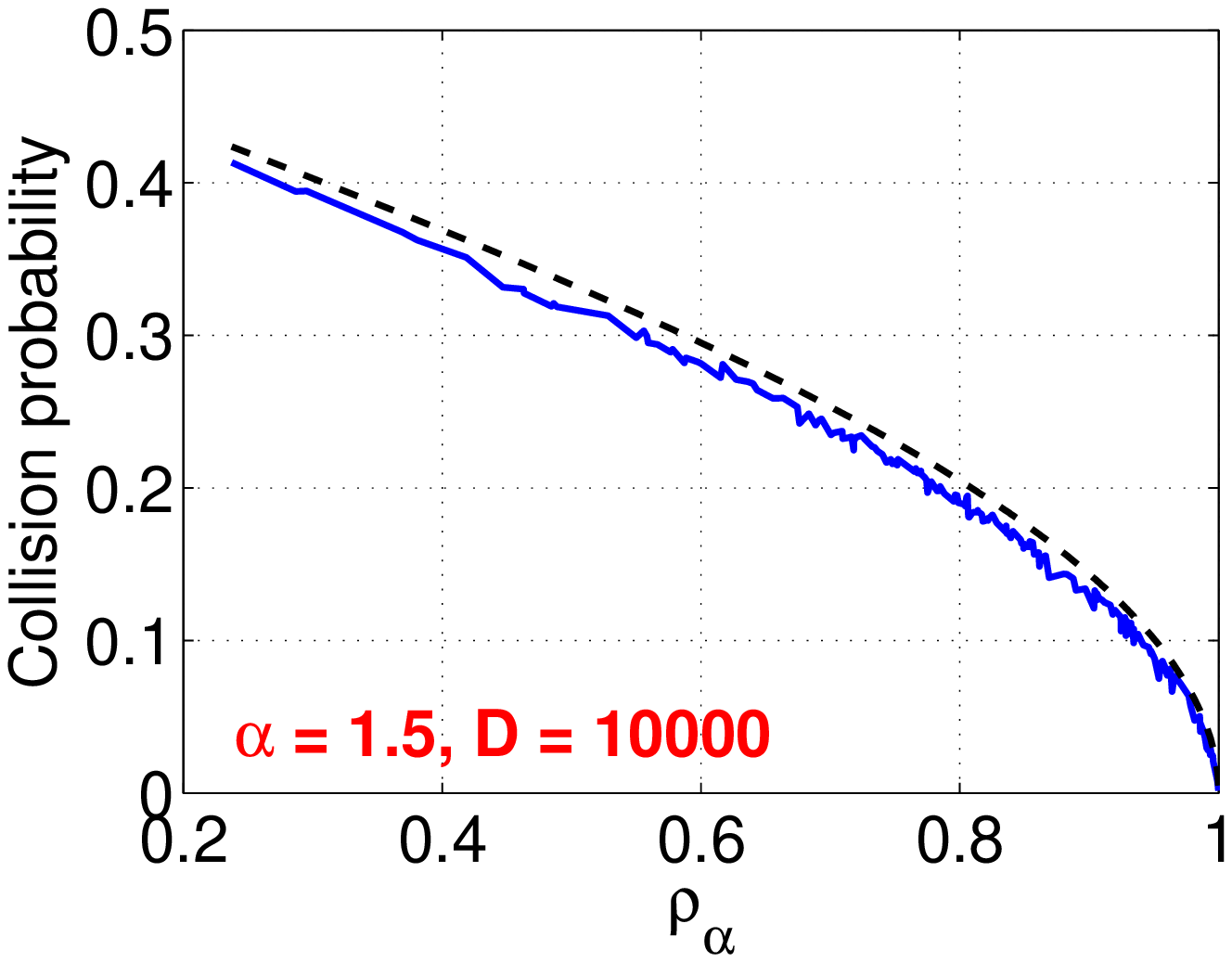}
\includegraphics[width=2.2in]{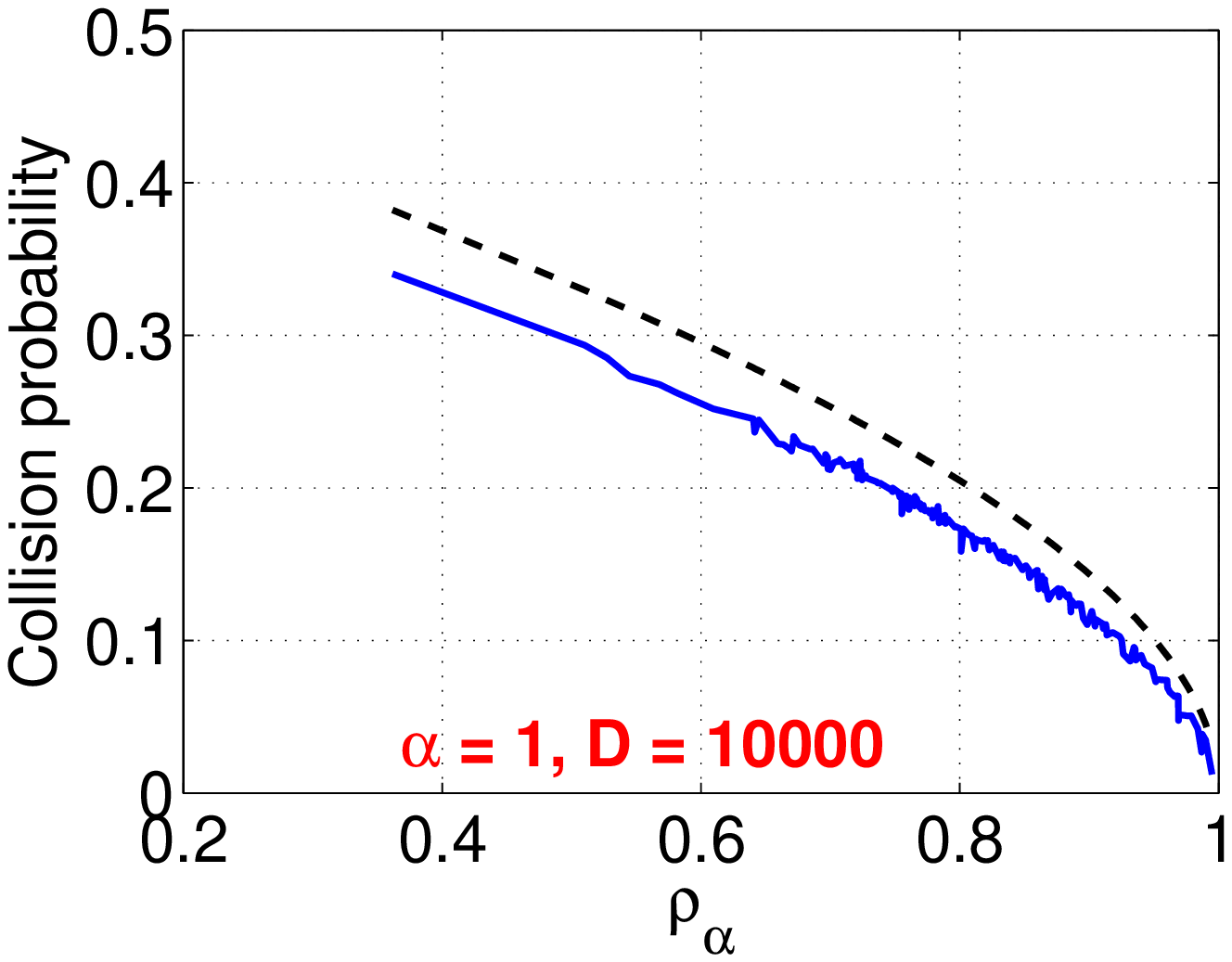}
\includegraphics[width=2.2in]{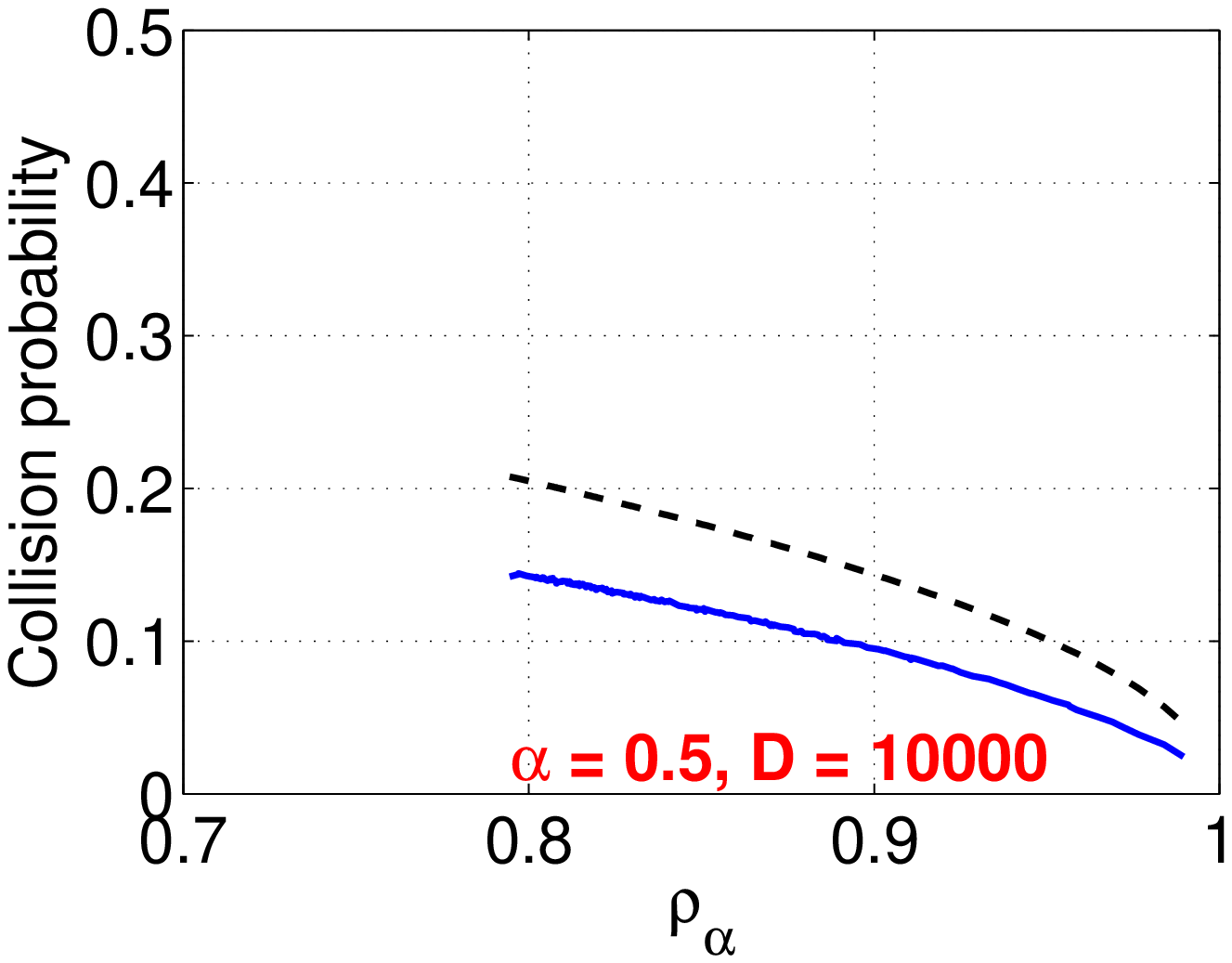}
}
\end{center}
\vspace{-0.2in}
\caption{\textbf{Dense Data} and $D=10,000$. Simulated collision probability $\mathbf{Pr}\left(\text{sign}(x) \neq \text{sign}(y)\right)$ for sign stable random projections. }\label{fig_RhoD10000T1}
\end{figure}

\begin{figure}[h!]
\begin{center}
\mbox{
\includegraphics[width=2.2in]{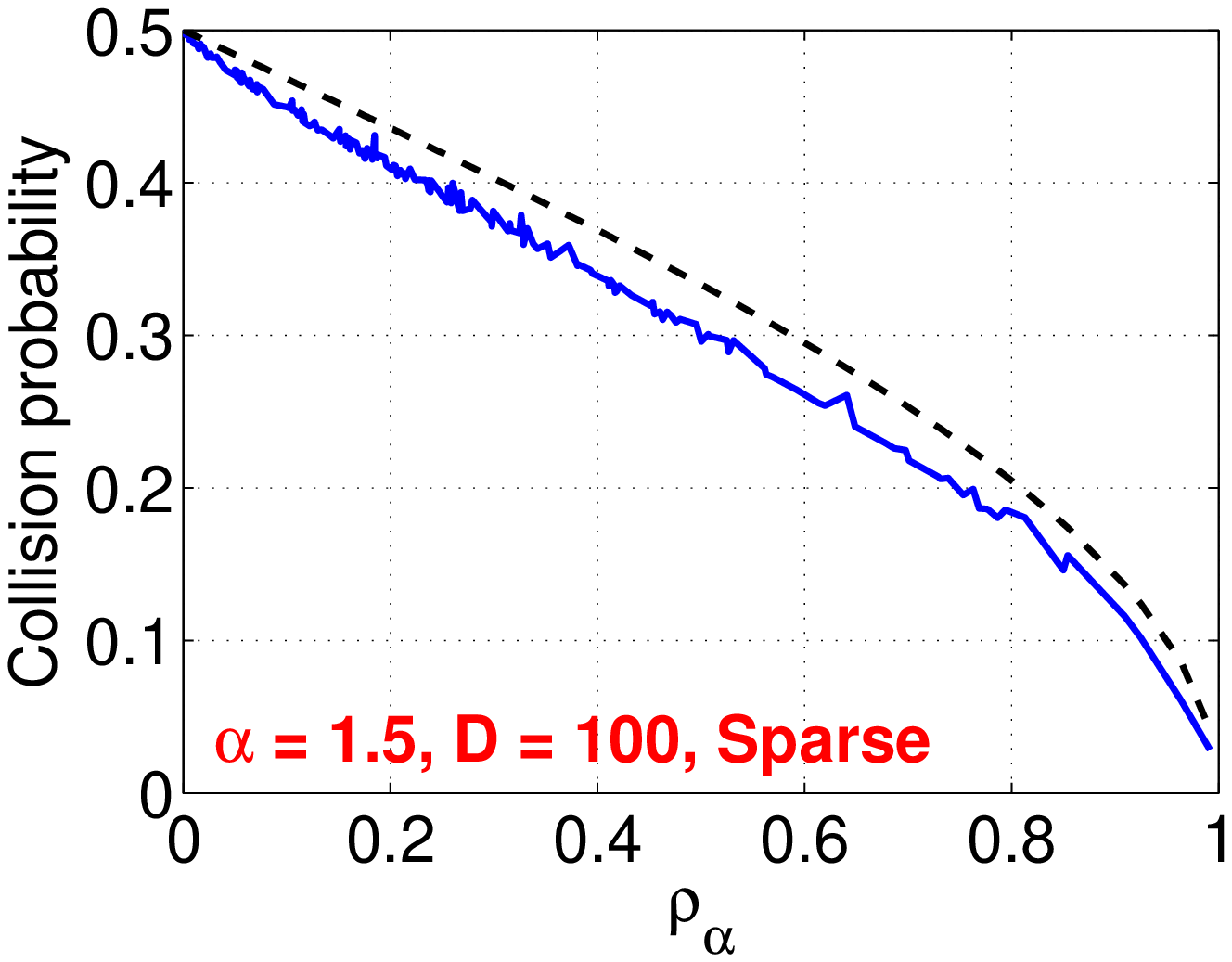}
\includegraphics[width=2.2in]{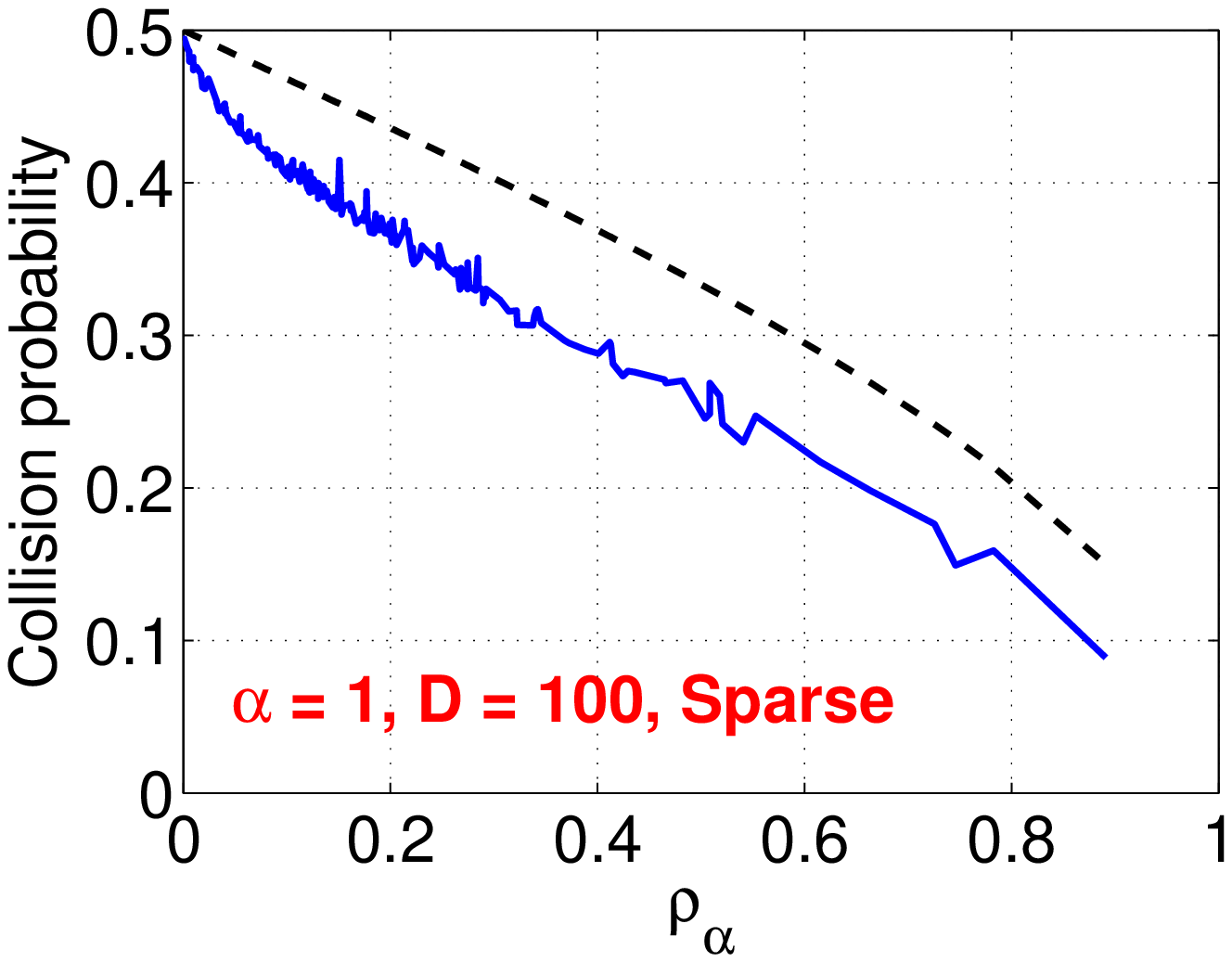}
\includegraphics[width=2.2in]{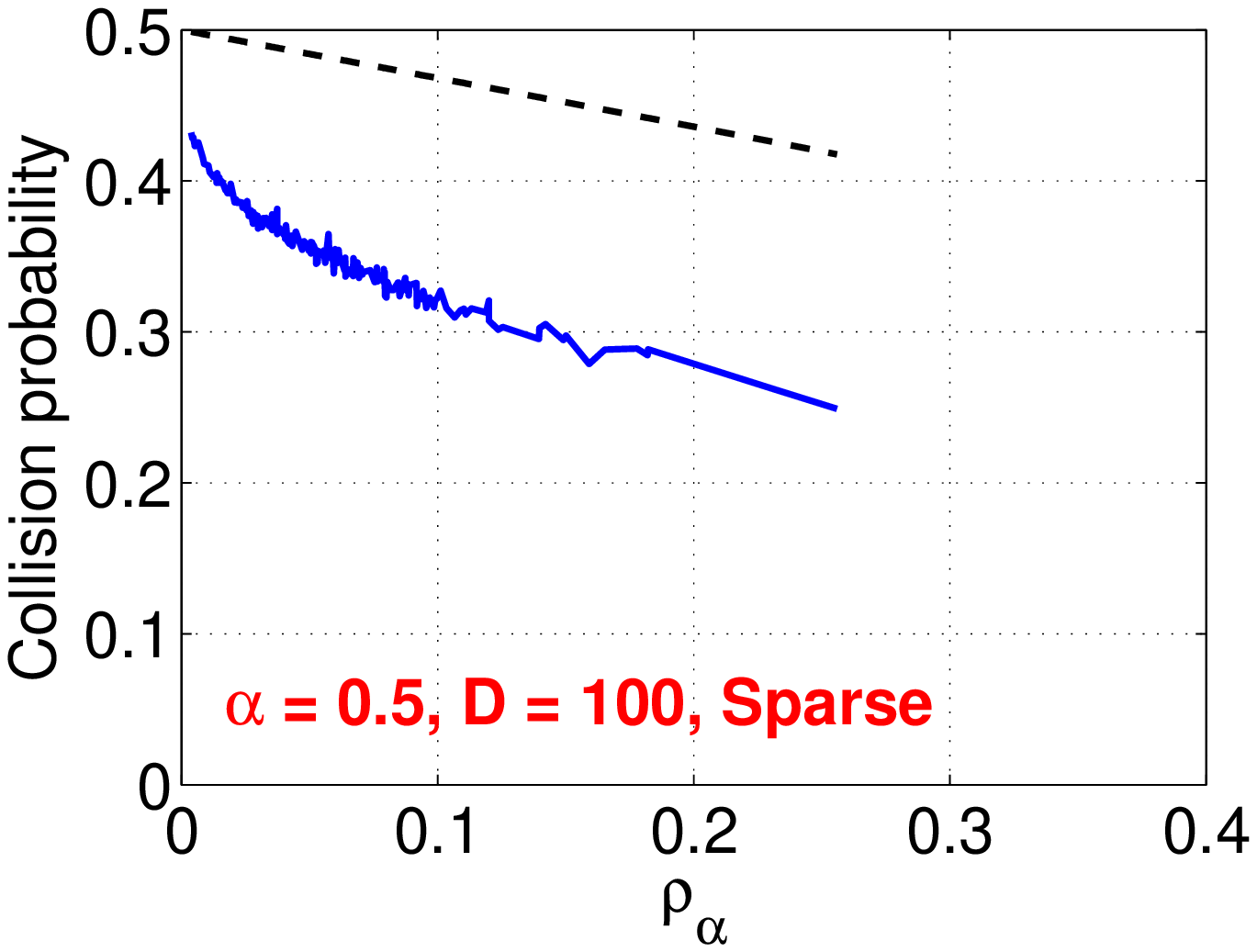}}
\mbox{

\includegraphics[width=2.2in]{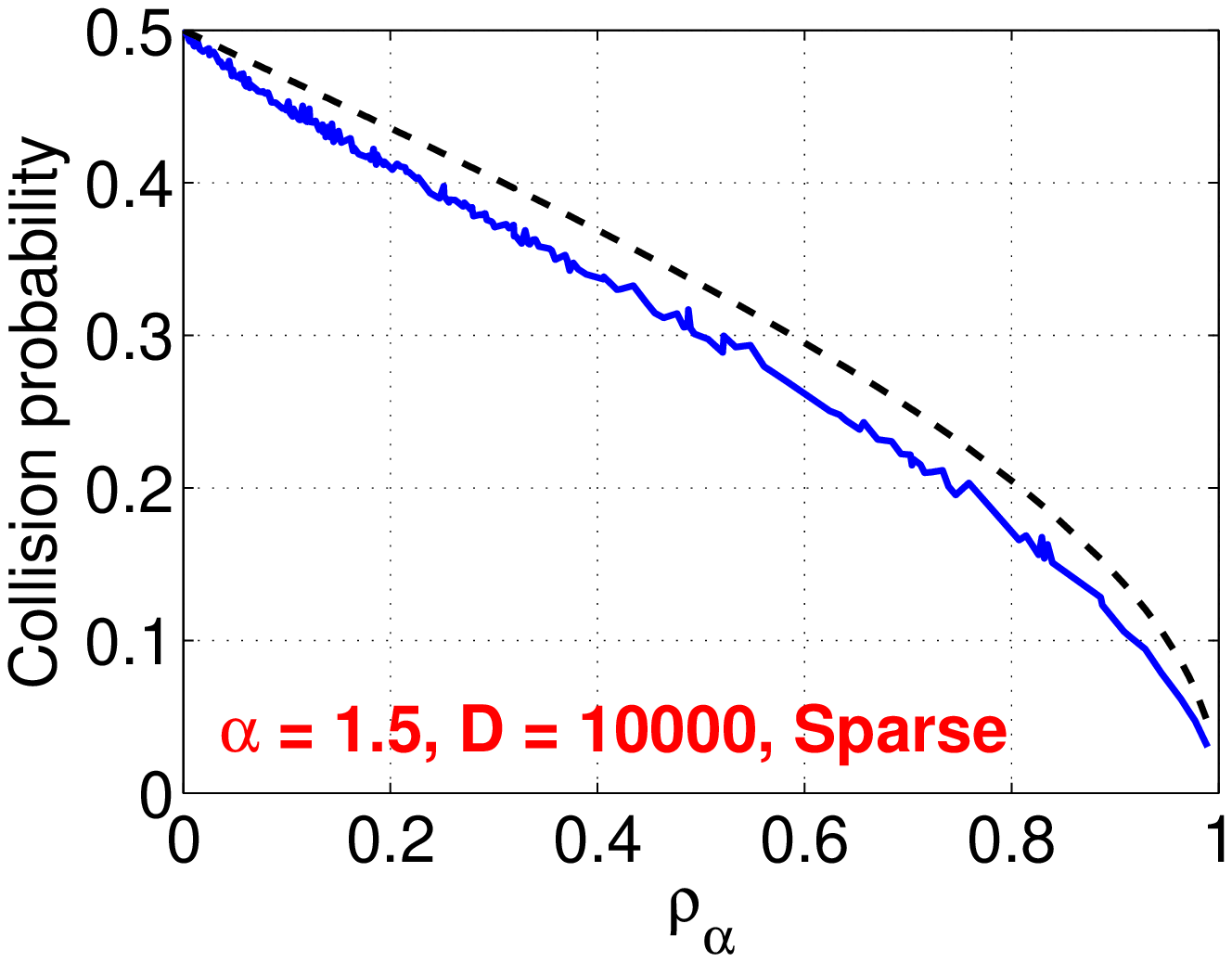}
\includegraphics[width=2.2in]{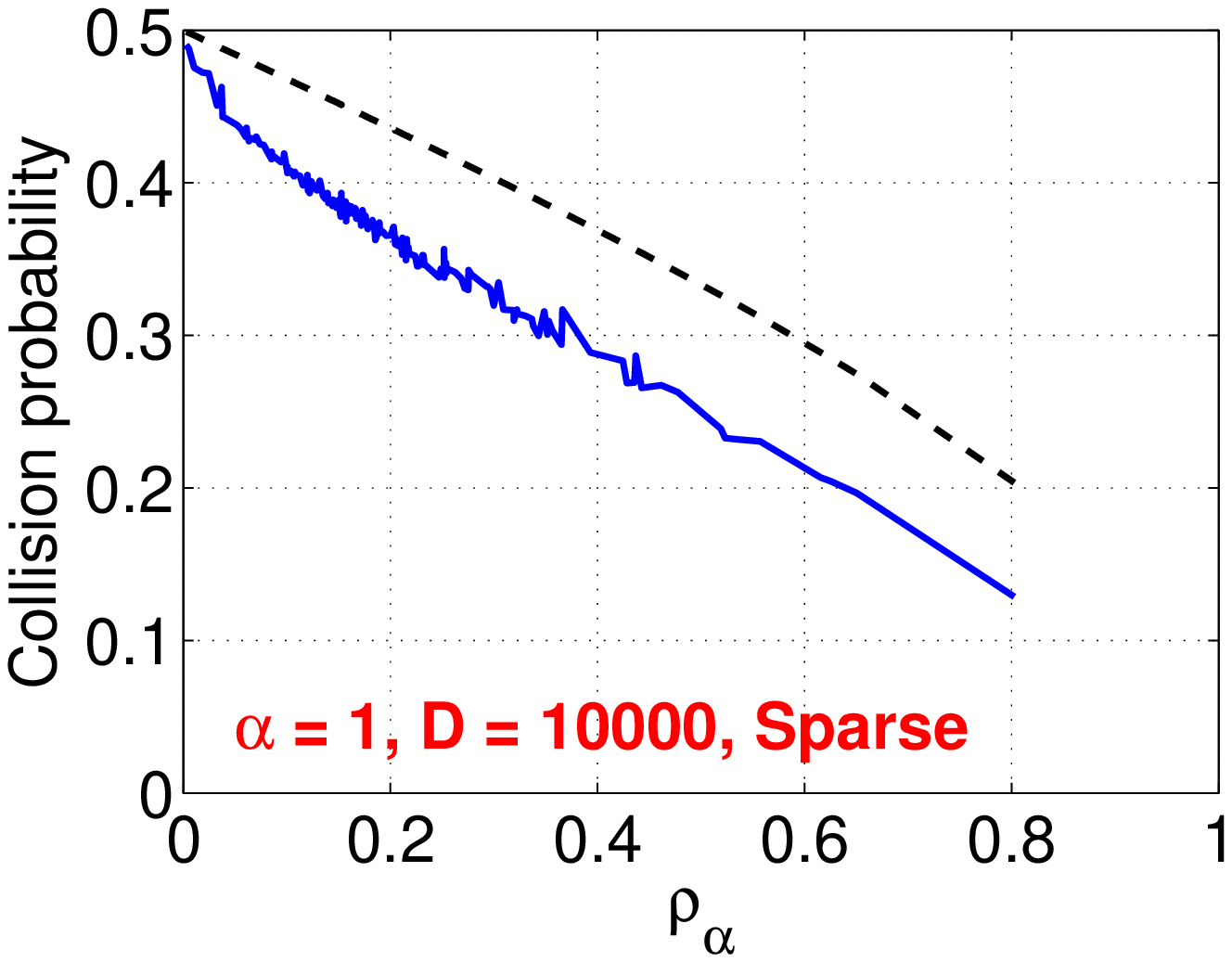}
\includegraphics[width=2.2in]{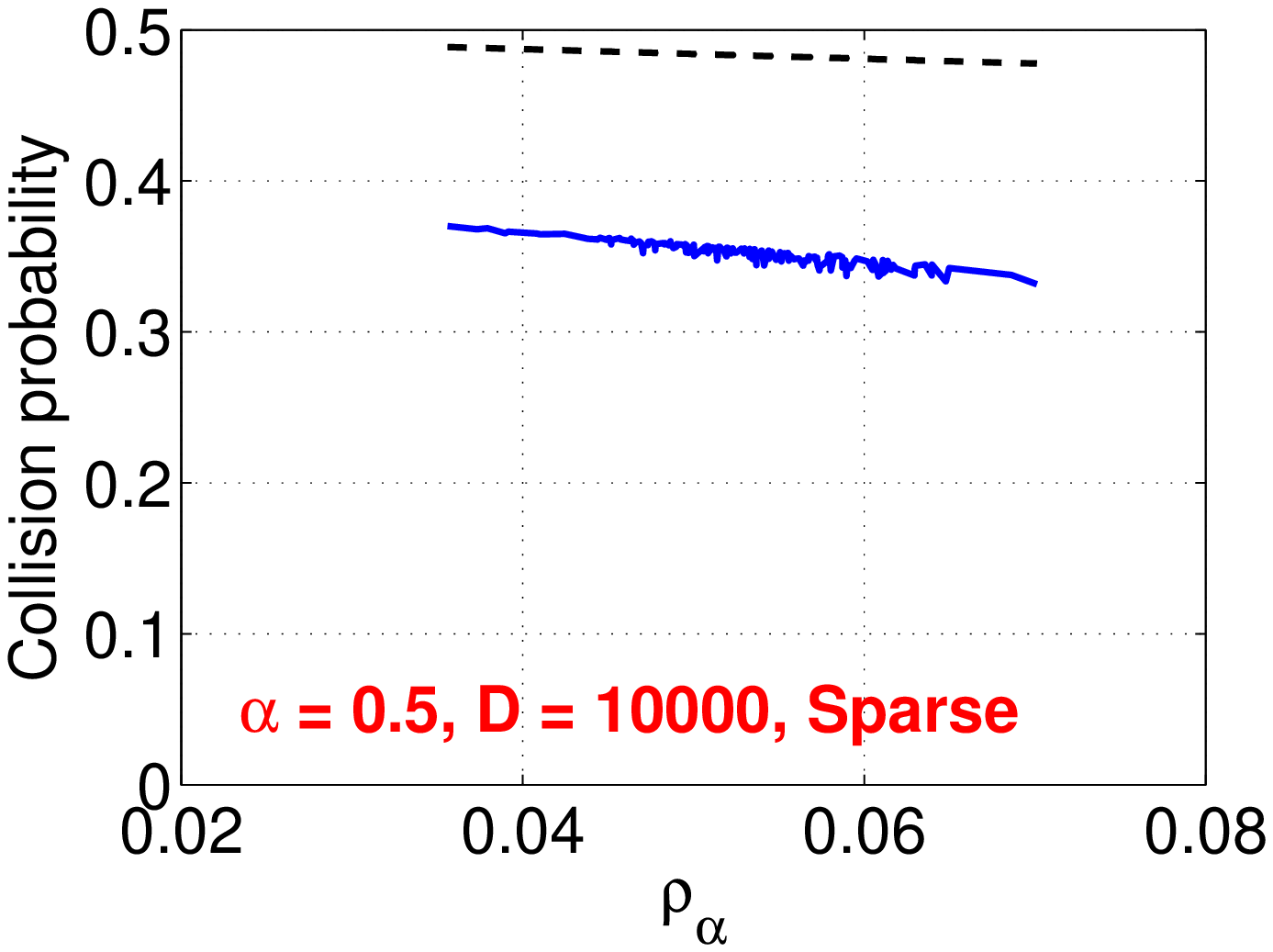}
}
\end{center}
\vspace{-0.2in}
\caption{\textbf{Sparse Data}. Simulated collision probability $\mathbf{Pr}\left(\text{sign}(x) \neq \text{sign}(y)\right)$ for sign stable random projection, for $D=100$ (upper panels)  and $D=10,000$ (bottom panels). The   upper bound is not  tight for $\alpha\leq1$. }\label{fig_RhoD100T1Sparse}
\end{figure}

Real-world high-dimensional datasets are often \textbf{sparse}. To verify the theoretical upper bound of the collision probability on sparse data, we  also simulate sparse data by randomly making $50\%$ of the generated data used in Figure~\ref{fig_RhoD100T1} be zero.  With sparse data, it is even more obvious that the theoretical upper bound $\frac{1}{\pi}\cos^{-1}\rho_\alpha$ is not sharp when $\alpha\leq1$, as shown in Figure~\ref{fig_RhoD100T1Sparse}.

In summary,  the   collision probability bound: $\mathbf{Pr}\left(\text{sign}(x) \neq \text{sign}(y)\right)\leq \frac{1}{\pi}\cos^{-1}\rho_\alpha$  is fairly sharp when $\alpha$ is close to 2 (e.g., $\alpha\geq 1.5$). However, for $\alpha\leq1$, a better approximation is needed.

\newpage\clearpage

\section{$\alpha =1$ and Chi-Square ($\chi^2$) Similarity}\label{sec_Cauchy}

In this section, we focus on nonnegative data ($u_i\geq 0, v_i\geq0$) and $\alpha=1$. This case is  important in practice. For example, we can view the data ($u_i$, $v_i$) as empirical probabilities, which are common when the data are generated from histograms (a popular technique in NLP and computer vision)~\cite{Article:Chapelle_99,Proc:Hein_AISTATS05,Proc:Jiang_CIVR07,Proc:Alexe_CVPR10,Proc:Wang_CVPR09,Proc:Vempati_BMVC10,Article:Vedaldi_PAMI12}.

In this context, we always normalize the data, i.e., $\sum_{i=1}^D u_i = \sum_{i=1}^D v_i =1$. From the  collision probability bound in Theorem~\ref{thm_upper_bound}, we know that
\begin{align}
&\mathbf{Pr}\left(\text{sign}(x) \neq \text{sign}(y)\right) \leq\frac{1}{\pi}\cos^{-1}\rho_1,\ \ \  \text{ where }  \\
&\rho_1 = \left(\frac{\sum_{i=1}^D u_i^{1/2}v_i^{1/2}}{\sqrt{\sum_{i=1}^Du_i\sum_{i=1}^D v_i}}\right)^{2} =\left(\sum_{i=1}^D u_i^{1/2}v_i^{1/2}\right)^{2}
\end{align}
This bound is not tight, as illustrated by our simulation study.  Interestingly, this collision probability can be related to the $\chi^2$ similarity.

Recall the definitions of the chi-square distance  $d_{\chi^2} = \sum_{i=1}^D \frac{(u_i-v_i)^2}{u_i+v_i}$  and the chi-square similarity $\rho_{\chi^2} = 1-\frac{1}{2}d_{\chi^2} = \sum_{i=1}^D\frac{2u_iv_i}{u_i+v_i}$. In this context, we should view $\frac{0}{0} = 0$.

\begin{lemma}
Assume $u_i\geq 0, v_i\geq 0$, $\sum_{i=1}^D u_i = 1$, $\sum_{i=1}^D v_i=1$. Then
\begin{align}
\rho_{\chi^2}=\sum_{i=1}^D\frac{2u_iv_i}{u_i+v_i} \geq \rho_1=\left(\sum_{i=1}^Du_i^{1/2}v_i^{1/2}\right)^2
\end{align}
\textbf{Proof:} \ By the Cauchy-Schwarz inequality
\begin{align}\notag
&\sum_{i=1}^D u_i^{1/2}v_i^{1/2}
= \sum_{i=1}^D \frac{u_i^{1/2}v_i^{1/2}}{[(u_i+v_i)/2]^{1/2}}
\left(\frac{u_i+v_i}{2}\right)^{1/2}\\\notag
\leq& \left( \sum_{i=1}^D \frac{2u_iv_i}{u_i+v_i}\right)^{1/2}
\left(  \sum_{i=1}^D\frac{u_i+v_i}{2}\right)^{1/2}
= \left( \sum_{i=1}^D \frac{2u_iv_i}{u_i+v_i}\right)^{1/2}\cdot 1
= \left( \sum_{i=1}^D \frac{2u_iv_i}{u_i+v_i}\right)^{1/2}
\end{align}$\hfill\square$
\end{lemma}

\vspace{0.1in}

It is know that the $\chi^2$-kernel $\rho_{\chi^2}$ is  PD~\cite{Proc:Hein_AISTATS05}. As a result, we know that the acos-$\chi^2$-kernel is  PD. The proof is analogous to the proof of Lemma~\ref{lem_rho2}.
\begin{lemma}
The kernel defined as $K(u,v) = 1-\frac{1}{\pi}\cos^{-1}\rho_{\chi^2}$ is positive definite (PD). \\

\noindent\textbf{Proof:}\  \ Because $\rho_{\chi^2}$ is PD, we can write it as an inner product $\rho_{\chi^2} = \sum_{i} \tilde{u}_i \tilde{v}_i$, even though we do not know $\tilde{u}_i$ and $\tilde{v}_i$ explicitly. When the two original vectors $u$ and $v$ are identical, we have $\rho_{\chi^2} =1$, which means we must have $\sum_{i} \tilde{u}_i^2 = \sum_{i} \tilde{v}_i^2 = 1$. At this point, we can apply normal random projections $x = \sum_i \tilde{u}_i r_i$,\ $y = \sum_i \tilde{v}_i r_i$, where $r_i \sim N(0,1)$, i.i.d. Then, just like Lemma~\ref{lem_rho2}, the desired result follows from the properties of indicator function and expectation. Note that there is also an equivalent argument in terms of a Gaussian process. Since $\rho_{\chi^2}$ is PD, there exists a Gaussian process with covariance  $\rho_{\chi^2}$, whose variance (i.e., when $u_i=v_i$) is exactly $1 = \sum_{i=1}^D u_i = \sum_{i=1}^D \frac{2u_iu_i}{u_i+u_i}$.
\hfill$\square$\\
\end{lemma}

The remaining question is how to connect {\em Cauchy random projections} with the $\chi^2$ similarity.

\section{Two Approximations of Collision Probability for Sign Cauchy  Projections}\label{sec_approximation}

\begin{figure}[h!]
\begin{center}
\mbox{
\includegraphics[width=2.8in]{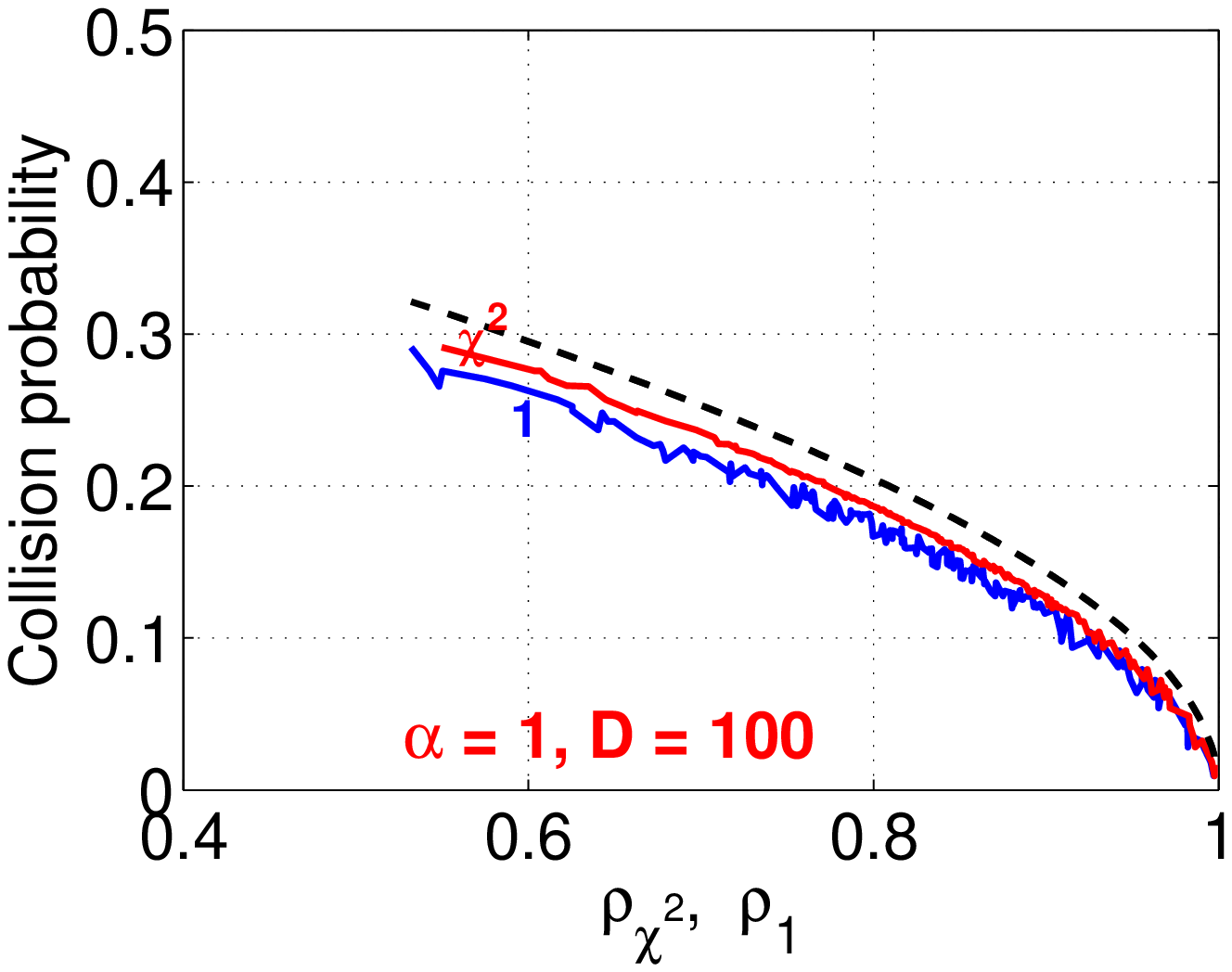}\hspace{0.3in}
\includegraphics[width=2.8in]{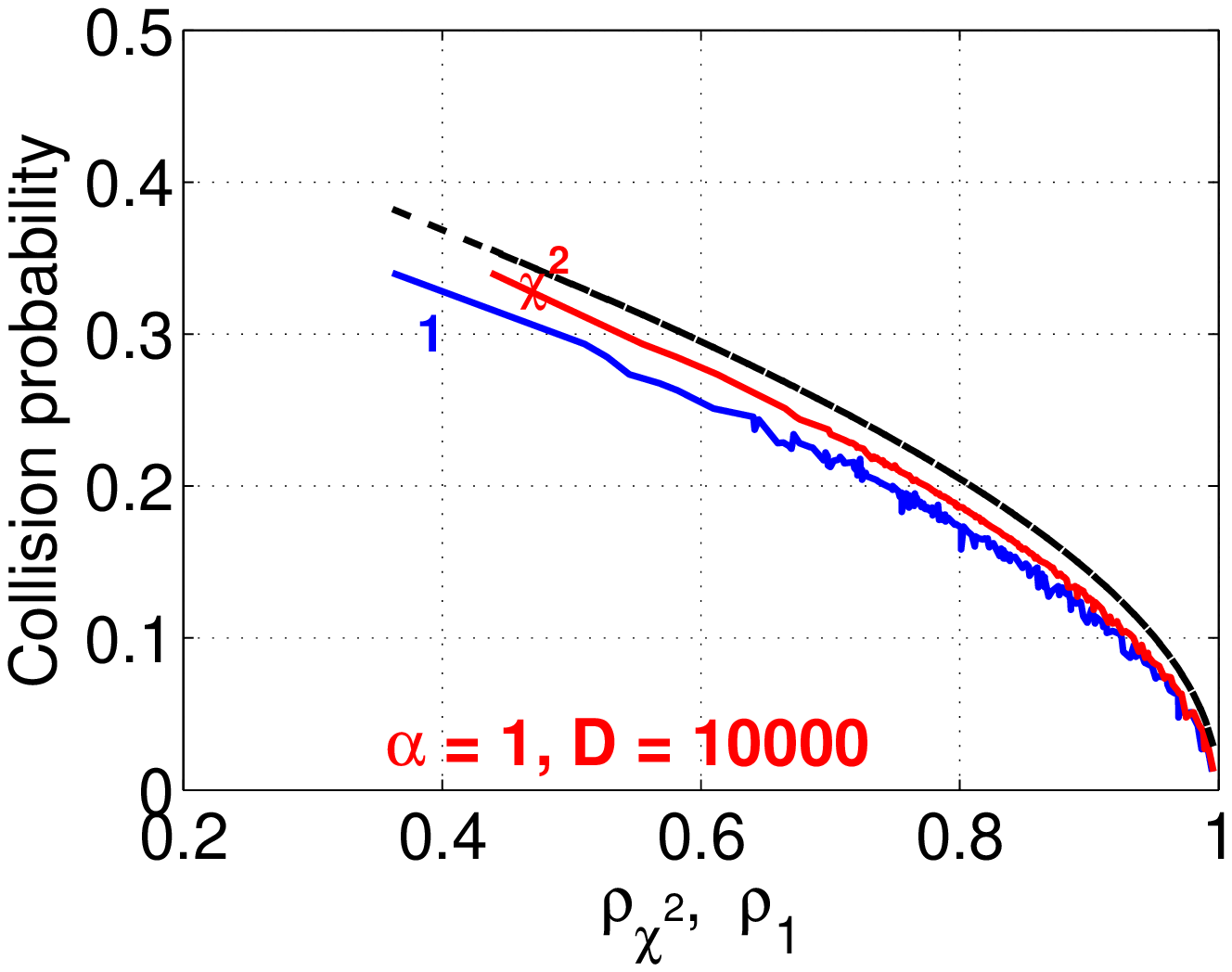}
}
\mbox{
\includegraphics[width=2.8in]{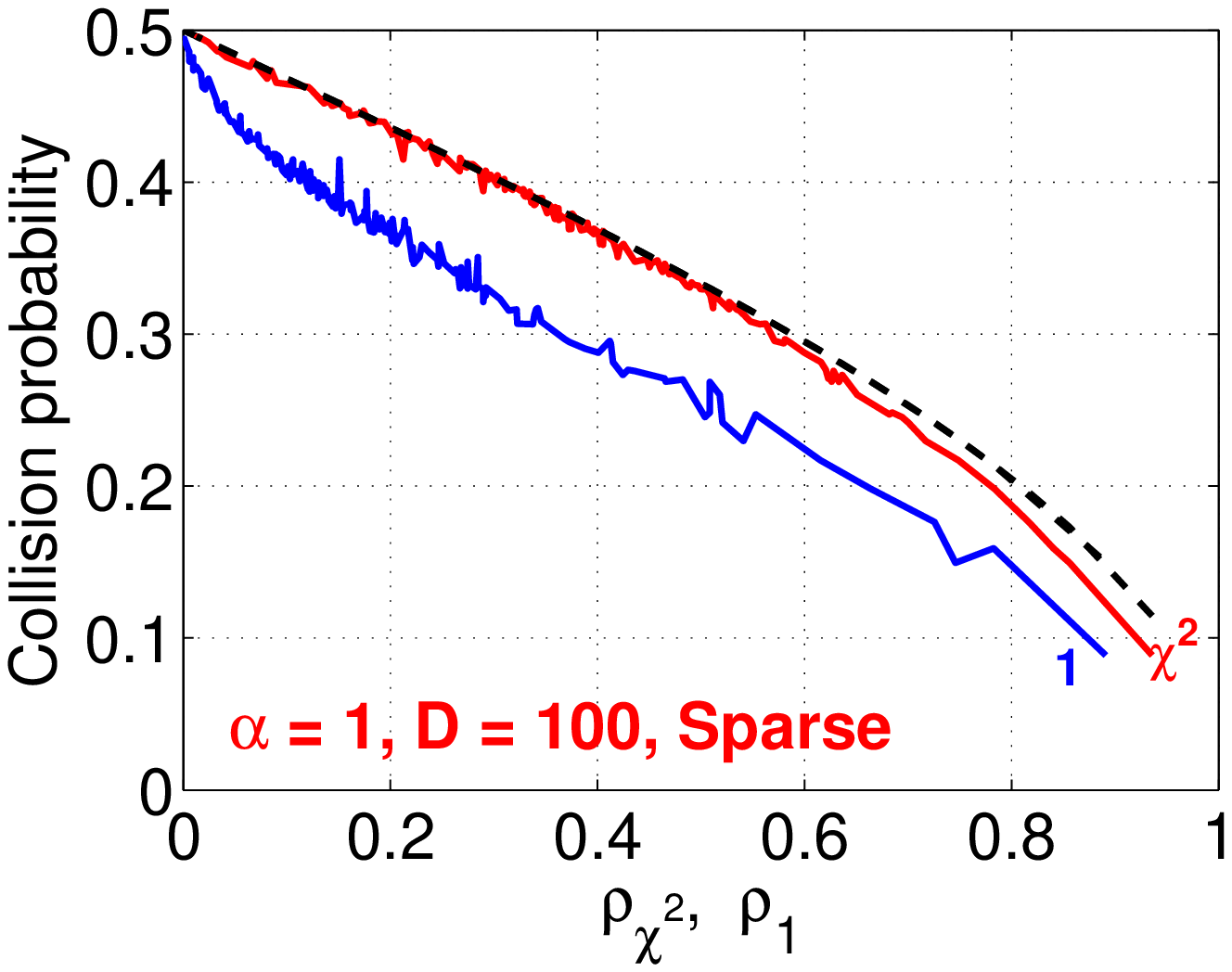}\hspace{0.3in}
\includegraphics[width=2.8in]{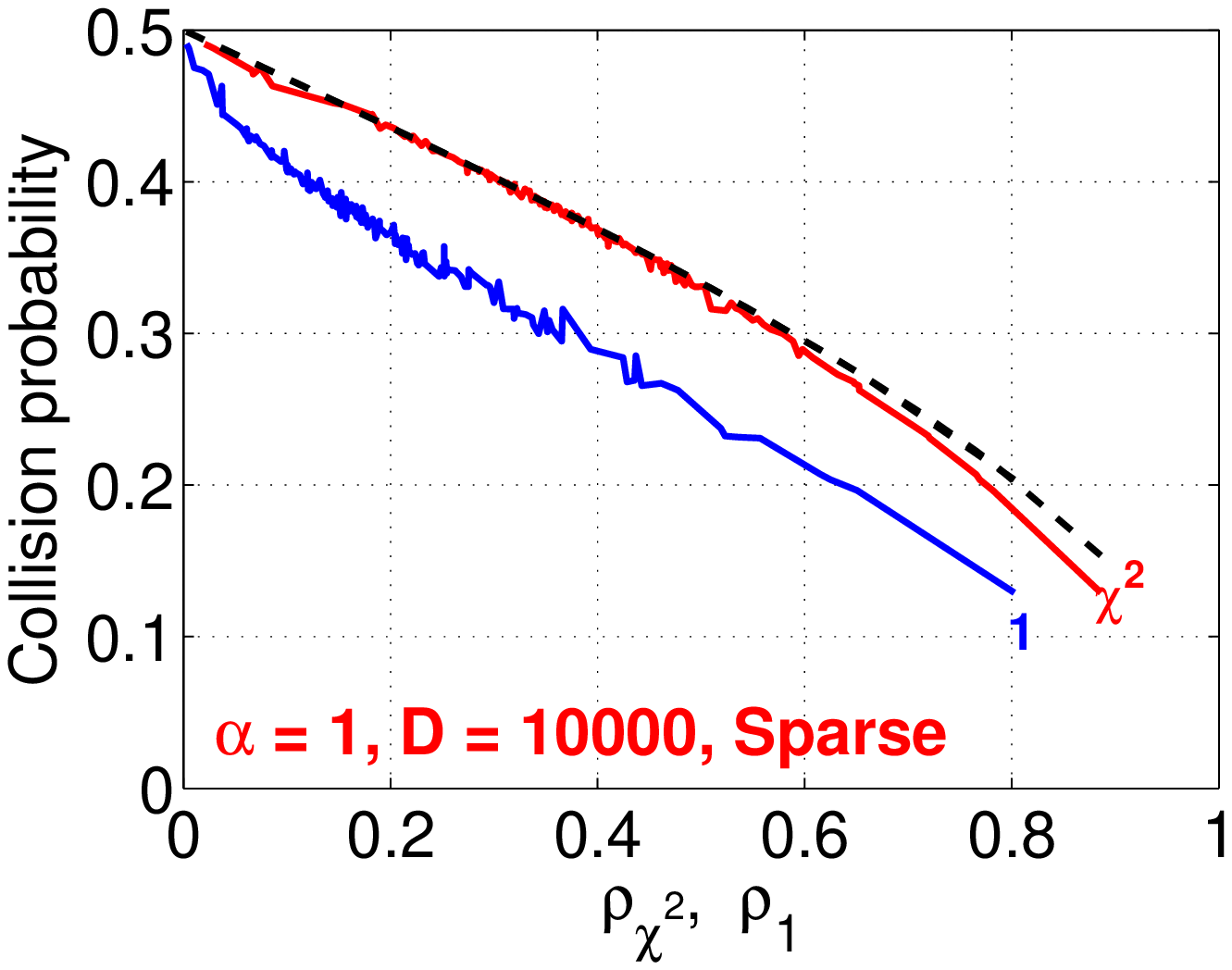}
}
\end{center}
\vspace{-0.2in}
\caption{The dashed curve is $\frac{1}{\pi}\cos^{-1}\left(\rho\right)$, where $\rho$ can be $\rho_1$ or $\rho_{\chi^2}$ depending on the context. In each panel, the two solid curves are the empirical collision probabilities in terms of $\rho_1$ (labeled by ``1'') or $\rho_{\chi^2}$ (labeled by ``$\chi^2$). It is clear that the proposed approximation $\frac{1}{\pi}\cos^{-1}\rho_{\chi^2}$ in (\ref{eqn_chi2(1)}) is more tight than the upper bound $\frac{1}{\pi}\cos^{-1}\rho_{1}$, especially so in sparse data. }\label{fig_RhoT1Cauchy1}
\end{figure}

It is a difficult  problem to derive the collision probability of sign Cauchy projections, especially if we would like to express the probability only in terms of certain summary statistics. Our \textbf{first} observation is that the collision probability can be well approximated using the $\chi^2$ similarity:
\begin{align}\label{eqn_chi2(1)}
\mathbf{Pr}\left(\text{sign}(x) \neq \text{sign}(y)\right) \approx  P_{\chi^2(1)}= \frac{1}{\pi}\cos^{-1}\left(\rho_{\chi^2}\right)
\end{align}
Figure~\ref{fig_RhoT1Cauchy1} shows this approximation is better than  $\frac{1}{\pi}\cos^{-1}\left(\rho_{1}\right)$. In  sparse data, the approximation is  very accurate while the  bound $\frac{1}{\pi}\cos^{-1}\left(\rho_{1}\right)$ is not sharp (and the curve is not smooth in $\rho_1$).

Our \textbf{second} (and  less obvious) approximation is the following integral:
\begin{align}\label{eqn_chi2(2)}
\mathbf{Pr}\left(\text{sign}(x) \neq \text{sign}(y)\right) \approx P_{\chi^2(2)} =\frac{1}{2}-\frac{2}{\pi^2}\int_0^{\pi/2} \tan^{-1}\left(\frac{\rho_{\chi^2}}{2-2\rho_{\chi^2}}\tan t\right) dt
\end{align}

Both approximations, $P_{\chi^2(1)}$ and $P_{\chi^2(2)}$, are monotone functions of $\rho_{\chi^2}$.  We can check the two boundary cases. When $\rho_{\chi^2} = 0$ (i.e., orthogonal data), we have $P_{\chi^2(1)}=P_{\chi^2(2)} = 1/2$ (i.e., the chance of collision is random). When  $\rho_{\chi^2} = 1$ (i.e., identical data), we have $P_{\chi^2(1)}=P_{\chi^2(2)} = 0$ (i.e.,  two signs always equal).

In practice (e.g., near neighbor search or classification), we normally do not need the $\rho_{\chi^2}$ values explicitly. It usually suffices if the collision probability is a monotone function of the similarity. For theoretical analysis, of course, it is desirable to know the explicit expressions.

\begin{figure}[h!]
\begin{center}
\mbox{
\includegraphics[width=2.8in]{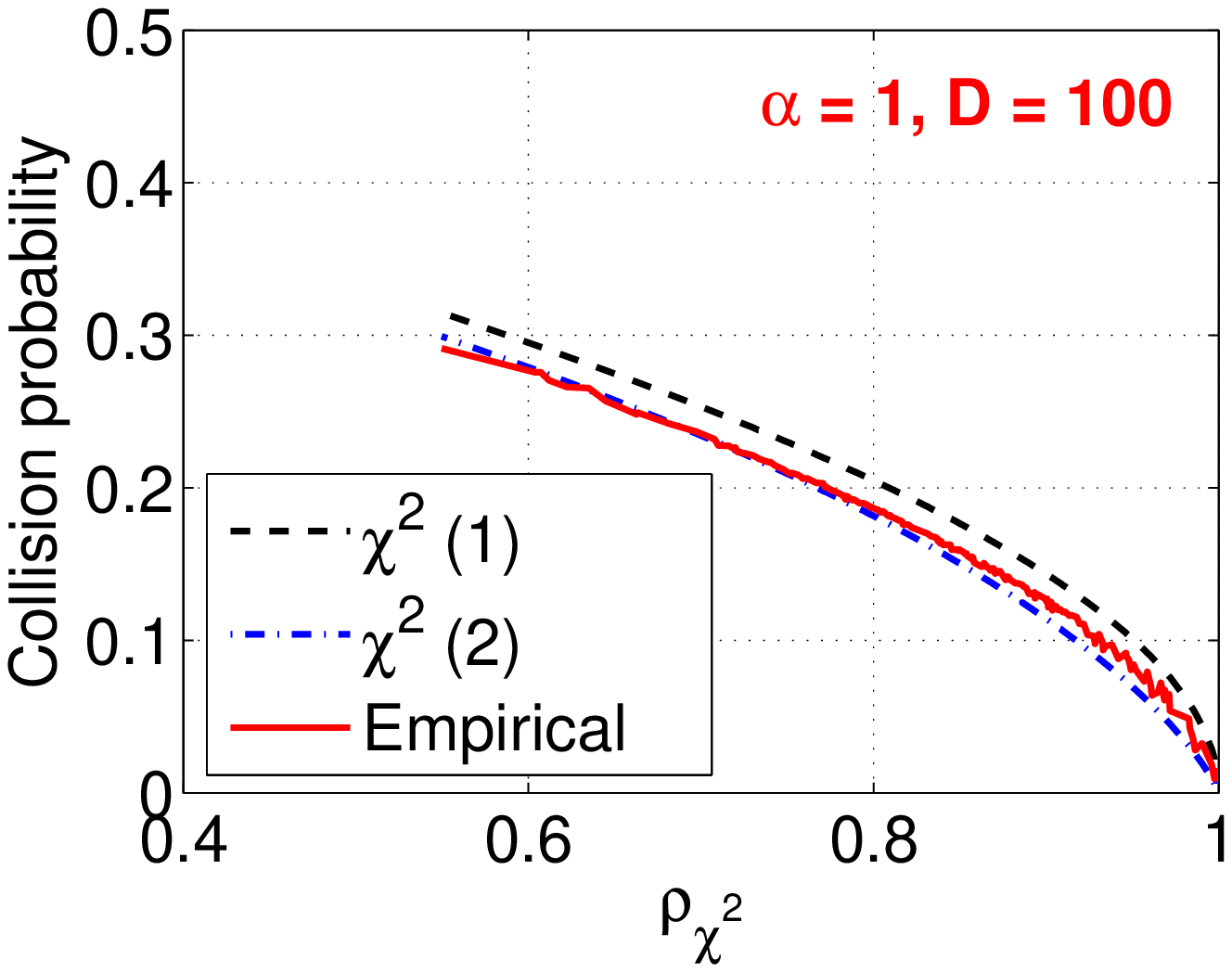}
\includegraphics[width=2.8in]{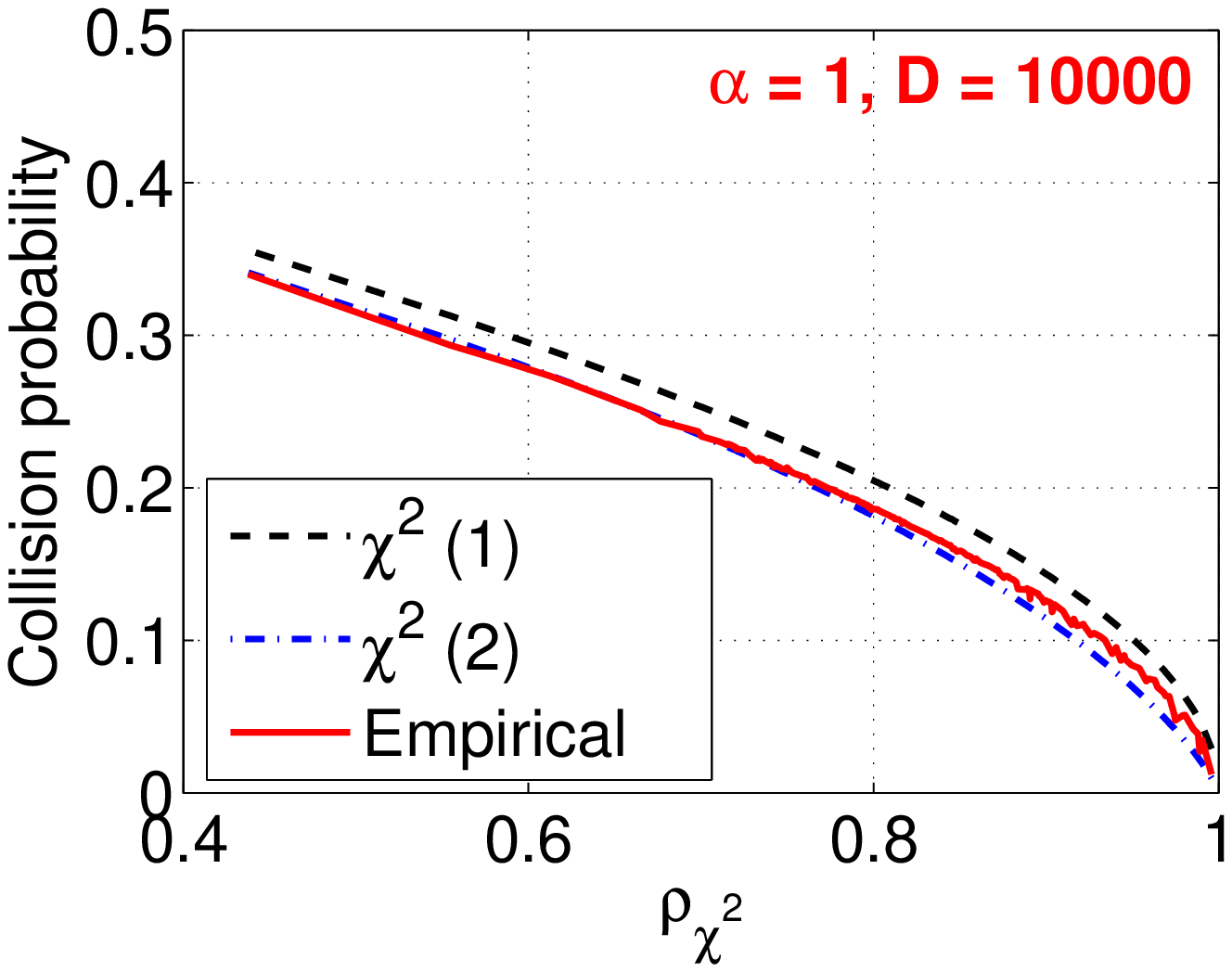}
}
\mbox{
\includegraphics[width=2.8in]{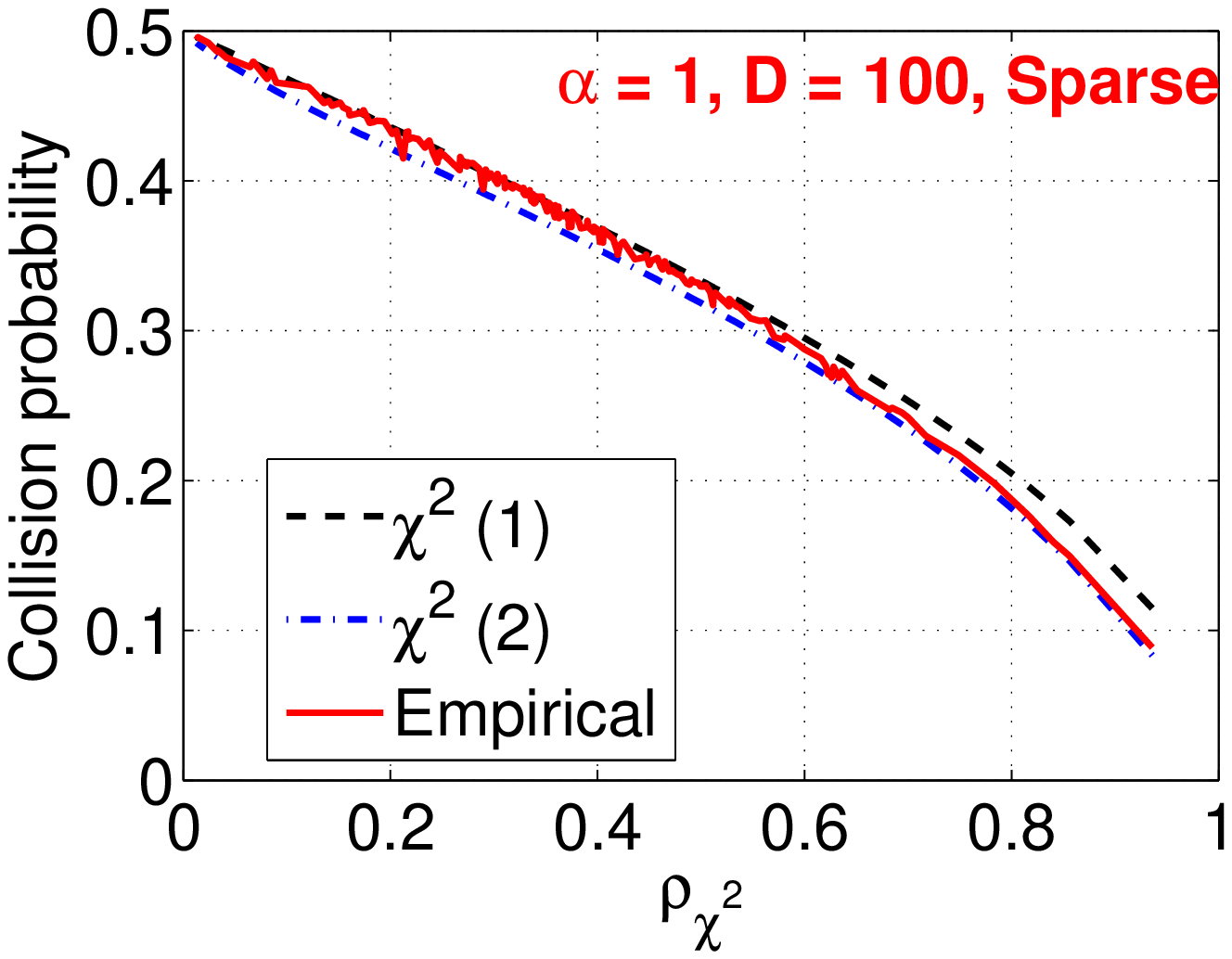}
\includegraphics[width=2.8in]{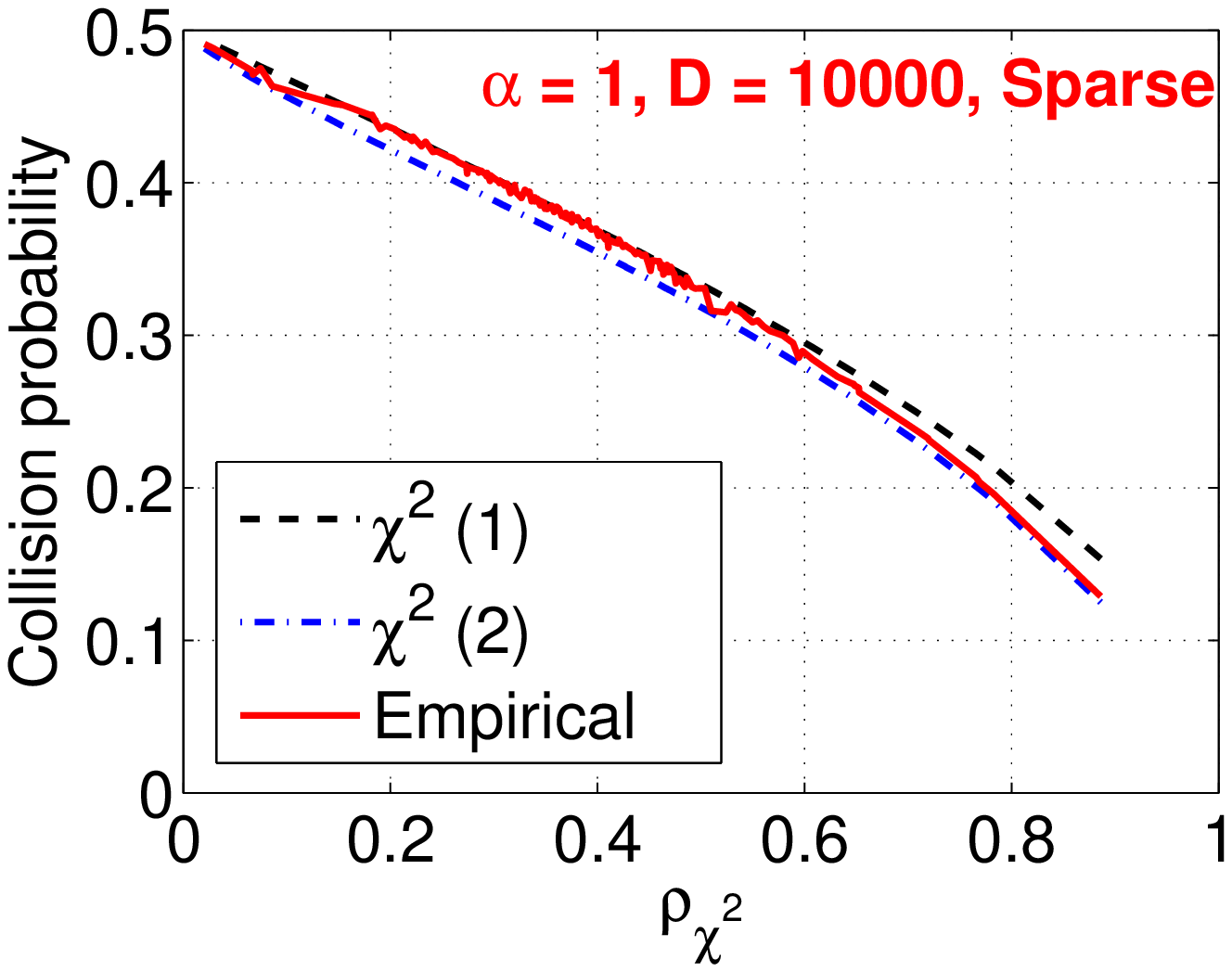}
}
\end{center}
\vspace{-0.2in}
\caption{Comparison of two approximations: $\chi^2 (1)$ based on (\ref{eqn_chi2(1)}) and $\chi^2 (2)$ based on (\ref{eqn_chi2(2)}). The solid curves (empirical probabilities expressed in terms of $\rho_{\chi^2}$) are the same solid curves labeled ``$\chi^2$'' in Figure~\ref{fig_RhoT1Cauchy1}. The top panels show that the second approximation (\ref{eqn_chi2(2)}) is  more accurate in dense data. The bottom panels illustrate that both  approximations are accurate in sparse data. (\ref{eqn_chi2(1)}) is slightly more accurate at small $\rho_{\chi^2}$ and (\ref{eqn_chi2(2)}) is more accurate at $\rho_{\chi^2}$ close to 1.}\label{fig_RhoT1Cauchy2}
\end{figure}

Figure~\ref{fig_RhoT1Cauchy2} illustrates that, for dense data, second approximation (\ref{eqn_chi2(2)}) is more accurate than the first approximation (\ref{eqn_chi2(1)}). The  second approximation  (\ref{eqn_chi2(2)}) is also accurate for sparse data in this set of simulations.

\begin{figure}[h!]
\begin{center}
\includegraphics[width=3in]{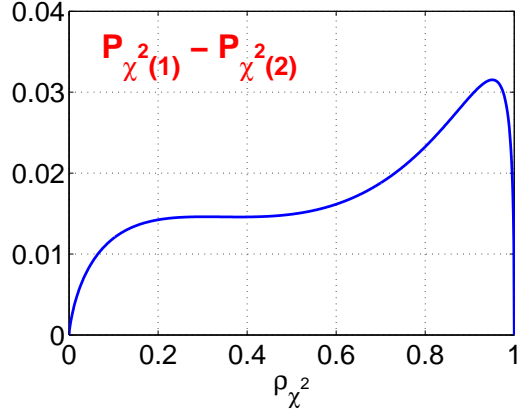}
\end{center}
\vspace{-0.2in}
\caption{The difference of the two $\chi^2$ approximations: $P_{\chi^2(1)}-P_{\chi^2(2)}$.}\label{fig_PrChi2Diff}
\end{figure}

Moreover, Figure~\ref{fig_PrChi2Diff} shows that $P_{\chi^2(2)}  \leq  P_{\chi^2(1)}$ (as proved in Lemma~\ref{lem_PrChi2Diff})

\begin{lemma}\label{lem_PrChi2Diff}
\begin{align}
P_{\chi^2(2)}  \leq  P_{\chi^2(1)}
\end{align}
\textbf{Proof:}\ \ For $0\leq z\leq 1$, to show
\begin{align}\notag
g(z) = \frac{1}{\pi}\cos^{-1}\left(z\right) + \frac{2}{\pi^2}\int_0^{\pi/2} \tan^{-1}\left(\frac{z}{2-2z}\tan t\right) dt - \frac{1}{2}\geq 0
\end{align}
Note that $g(0)=0$, $g(1) =0$, and
\begin{align}\notag
g^\prime(z) =& -\frac{1}{\pi}\frac{1}{\sqrt{1-z^2}} +\frac{1}{\pi^2}\int_0^{\pi/2}\frac{\frac{1}{(1-z)^2}\tan t}{1+\left(\frac{z}{2-2z}\tan t\right)^2}  dt\\\notag
=&-\frac{1}{\pi}\frac{1}{\sqrt{1-z^2}} + \frac{2}{\pi^2}\frac{1}{\left(z^2-4(1-z)^2\right)}\log \frac{z^2}{4(1-z)^2}
\end{align}
There are three extremum points (where $g^\prime(z)=0$), at $z_1 = 0.30555$, $z_2 = 0.38291$,  and $z_3 = 0.95094$ (keeping five digits). We can check $g(z)>0$ at all three extremum points. This completes the proof.\hfill$\Box$
\end{lemma}

\subsection{Binary Data}

Interestingly, when the data are binary (before normalization), we can compute the collision probability exactly, which allows us to analytically assess the accuracy of the approximations. In fact, this case inspired us to propose the second approximation (\ref{eqn_chi2(2)}), which is otherwise not intuitive.

For convenience, we define $a = |I_a|, \ \ b = |I_b|, \ \ c = |I_{c}|$, where
\begin{align}
I_a = \{i|u_i>0, v_i=0\},\hspace{0.2in} I_b = \{i|v_i>0, u_i=0\},\hspace{0.2in} I_{c} = \{i|u_i>0, v_i>0\},
\end{align}

Assume binary data (before normalization, i.e., sum to one). That is,
\begin{align}
u_i = \frac{1}{|I_a|+|I_{c}|}=\frac{1}{a+c}, \ \ \forall i\in I_a\cup I_c,\hspace{0.5in}
v_i = \frac{1}{|I_b|+|I_{c}|}=\frac{1}{b+c}, \ \ \forall i\in I_b\cup I_c
\end{align}
The chi-square similarity $\rho_{\chi^2}$  becomes
$\rho_{\chi^2} = \sum_{i=1}^D \frac{2u_iv_i}{u_i+v_i} =\frac{2c}{a+b+2c}$ and hence
$\frac{\rho_{\chi^2}}{2-2\rho_{\chi^2}} = \frac{c}{a+b}$.

\begin{theorem}\label{thm_binary_prob}
Assume binary data. When $\alpha=1$, the exact collision probability is
\begin{align}\label{eqn_binary_prob}
\mathbf{Pr}\left(\text{sign}(x)\neq \text{sign}(y)\right)
=&\frac{1}{2}-\frac{2}{\pi^2}E\left\{\tan^{-1}\left(\frac{c}{a}|R|\right)\tan^{-1}\left(\frac{c}{b}|R|\right)\right\}
\end{align}
where $R$ is a standard Cauchy random variable.\\

\noindent\textbf{Proof:}\ \ See Appendix~\ref{app_thm_binary_prob}.\hfill$\Box$\\
\end{theorem}

When $a=0$ or $b=0$,  we have $E\left\{\tan^{-1}\left(\frac{c}{a}|R|\right)\tan^{-1}\left(\frac{c}{b}|R|\right)\right\} = \frac{\pi}{2}E\left\{\tan^{-1}\left(\frac{c}{a+b}|R|\right)\right\}$. This observation inspires  us to propose the approximation (\ref{eqn_chi2(2)}):
\begin{align}\notag
P_{\chi^2(2)} = \frac{1}{2} - \frac{1}{\pi}  E\left\{\tan^{-1}\left(\frac{c}{a+b}|R|\right)\right\} = \frac{1}{2} - \frac{2}{\pi^2} \int_0^{\pi/2} \tan^{-1}\left(\frac{c}{a+b} \tan t\right) dt
\end{align}
To validate this approximation for binary data, we study the difference between (\ref{eqn_binary_prob}) and ({\ref{eqn_chi2(2)}), i.e.,
\begin{align}\notag
&Err  =  \mathbf{Pr}\left(\text{sign}(x)\neq \text{sign}(y)\right) - P_{\chi^2(2)}= Z(a/c,b/c)\\\label{eqn_ErrBinaryData}
=&
-\frac{2}{\pi^2}E\left\{\tan^{-1}\left(\frac{1}{a/c}|R|\right)\tan^{-1}\left(\frac{1}{b/c}|R|\right)\right\}
+\frac{1}{\pi} E\left\{\tan^{-1}\left(\frac{1}{a/c+b/c}|R|\right)\right\}
\end{align}
(\ref{eqn_ErrBinaryData}) can be easily computed by simulations. Figure~\ref{fig_ErrBinaryData} confirms that the errors are very small and larger than zero. The maximum error is smaller than 0.0192.
\begin{figure}[h!]
\begin{center}
\includegraphics[width=2.8in]{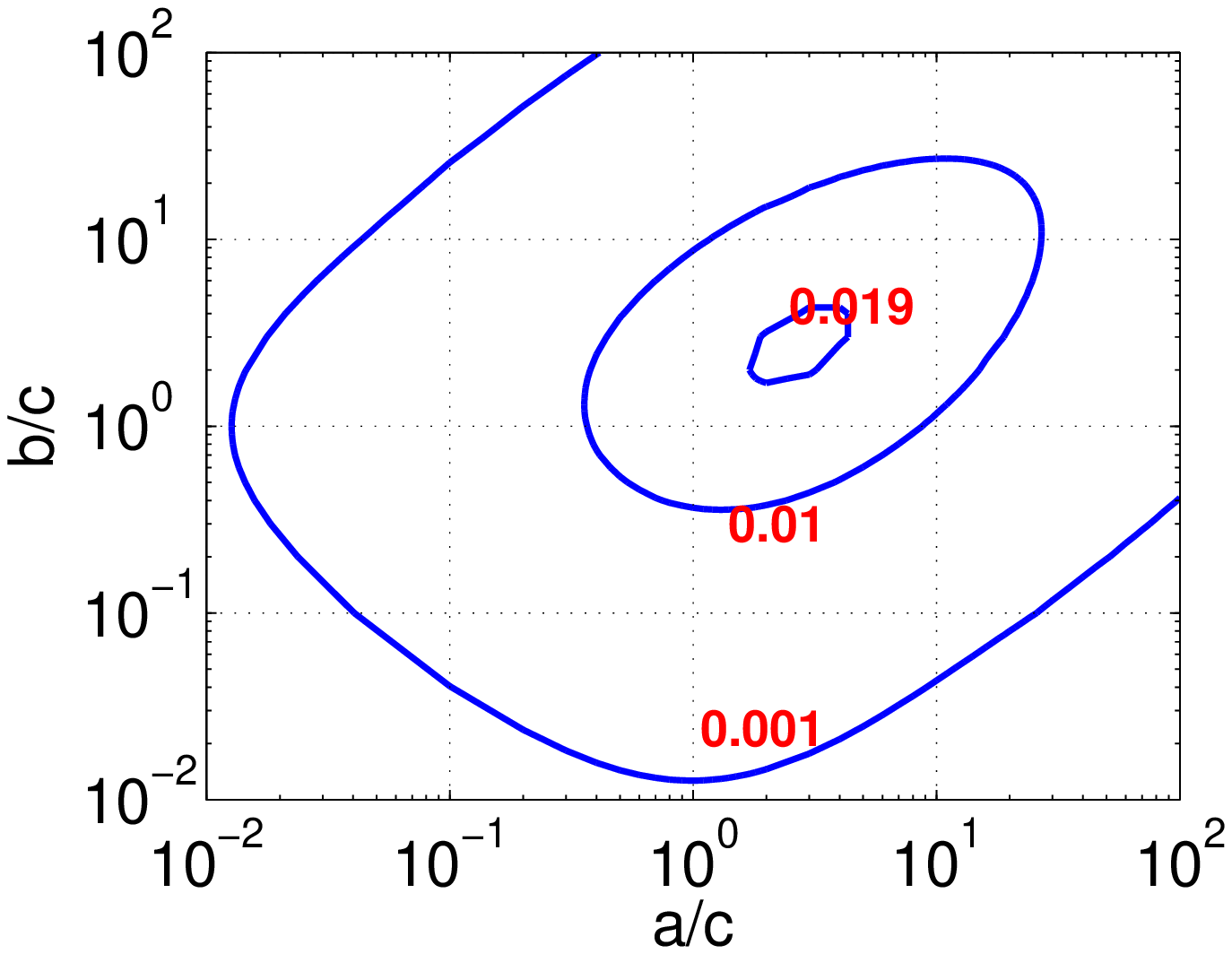}\hspace{0.3in}
\includegraphics[width=2.8in]{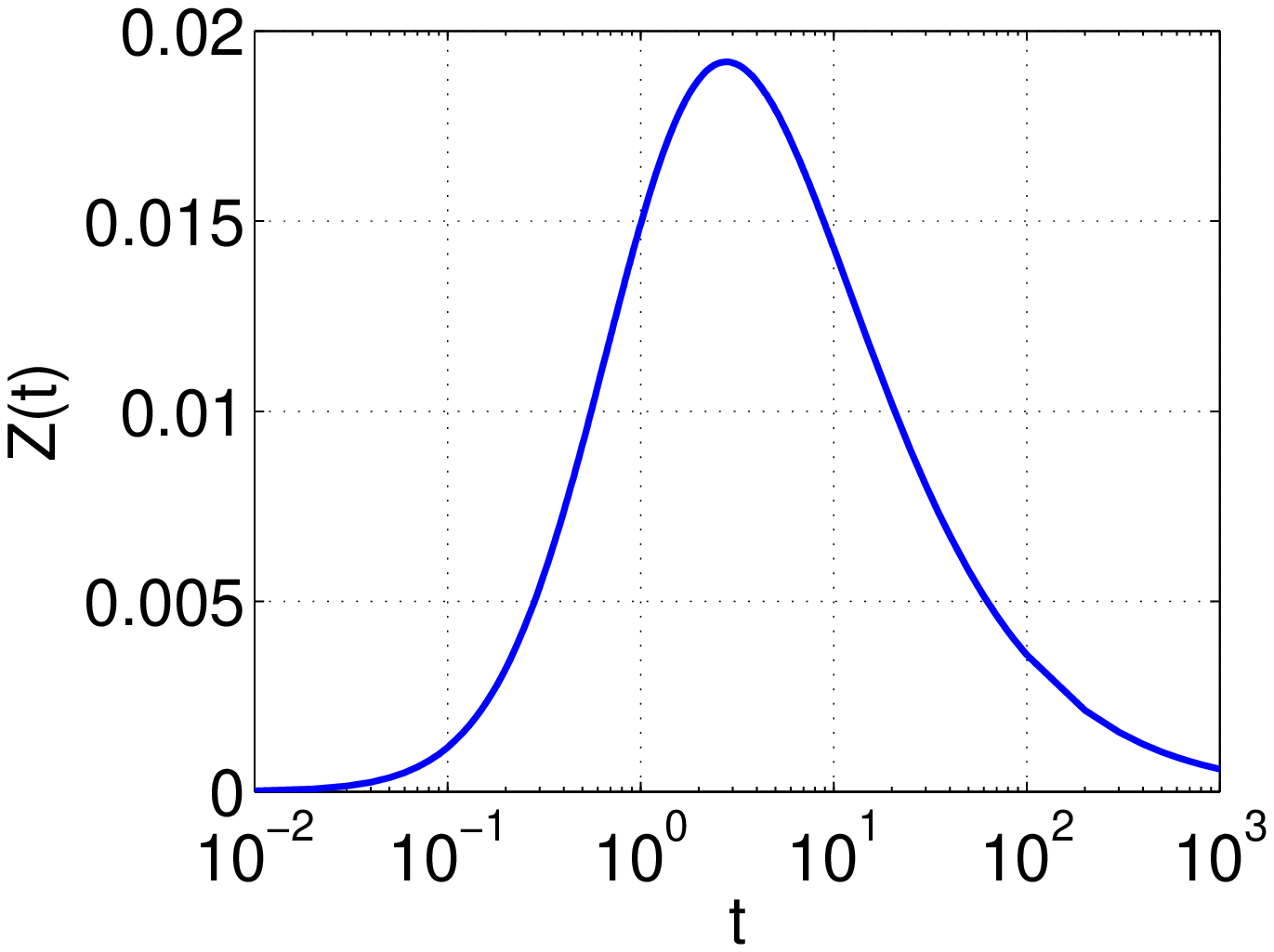}
\end{center}
\vspace{-0.2in}
\caption{Left panel: contour plot for the error  $Z(a/c,b/c)$  in (\ref{eqn_ErrBinaryData}). The maximum error (which is $<0.0192$) occurs along the diagonal line.  Right panel: the diagonal curve of $Z(a/c,b/c)$.
}\label{fig_ErrBinaryData}
\end{figure}

More precisely, we have the following Lemma.
\begin{lemma}\label{lem_ErrBinary}
The error defined in (\ref{eqn_ErrBinaryData}) ranges between 0 and $Z(t^*)$:
\begin{align}
0\leq Z(a/c,b/c) \leq Z(t^*)
=\int_0^{\infty} \left\{-\frac{2}{\pi^2}\left(\tan^{-1}\left(\frac{r}{t^*}\right)\right)^2 +\frac{1}{\pi}\tan^{-1}\left(\frac{r}{2t^*}\right)\right\} \frac{2}{\pi}\frac{1}{1+r^2}dr
\end{align}
where $t^* = 2.77935$ is the solution to $\frac{1}{t^2-1}\log\frac{2t}{1+t} - \frac{\log(2t)}{(2t)^2-1}=0$.  Numerically, $Z(t^*) = 0.01919$.

\noindent\textbf{Proof:}\ \ See Appendix~\ref{app_lem_ErrBinary}. We keep five digits for the numerical values. \hfill$\Box$
\end{lemma}

\subsection{An Experiment Based on 3.6 Million English Word Pairs}

To further validate the two $\chi^2$ approximations, we experiment with a word occurrences dataset (which is an example of histogram data) from a chunk of $D=2^{16}$ web crawl documents. There are in total 2,702 words, i.e., 2,702 vectors and 3,649,051 word pairs. The entries of a vector are the occurrences of the word. This is a typical sparse, non-binary dataset. Interestingly, the errors of the collision probabilities based on two $\chi^2$ approximations are still very small. To report the results, we apply sign Cauchy random projections $10^7$ times to evaluate the approximation errors of (\ref{eqn_chi2(1)}) and (\ref{eqn_chi2(2)}).

The results, as presented in Figure~\ref{fig_WordProb}, again confirm that the upper bound $\frac{1}{\pi}\cos^{-1}\rho_1$ is not tight and both $\chi^2$ approximations,  $P_{\chi_2(1)}$ and $P_{\chi_2(2)}$, are accurate.

\begin{figure}[h!]
\begin{center}
\includegraphics[width=2.8in]{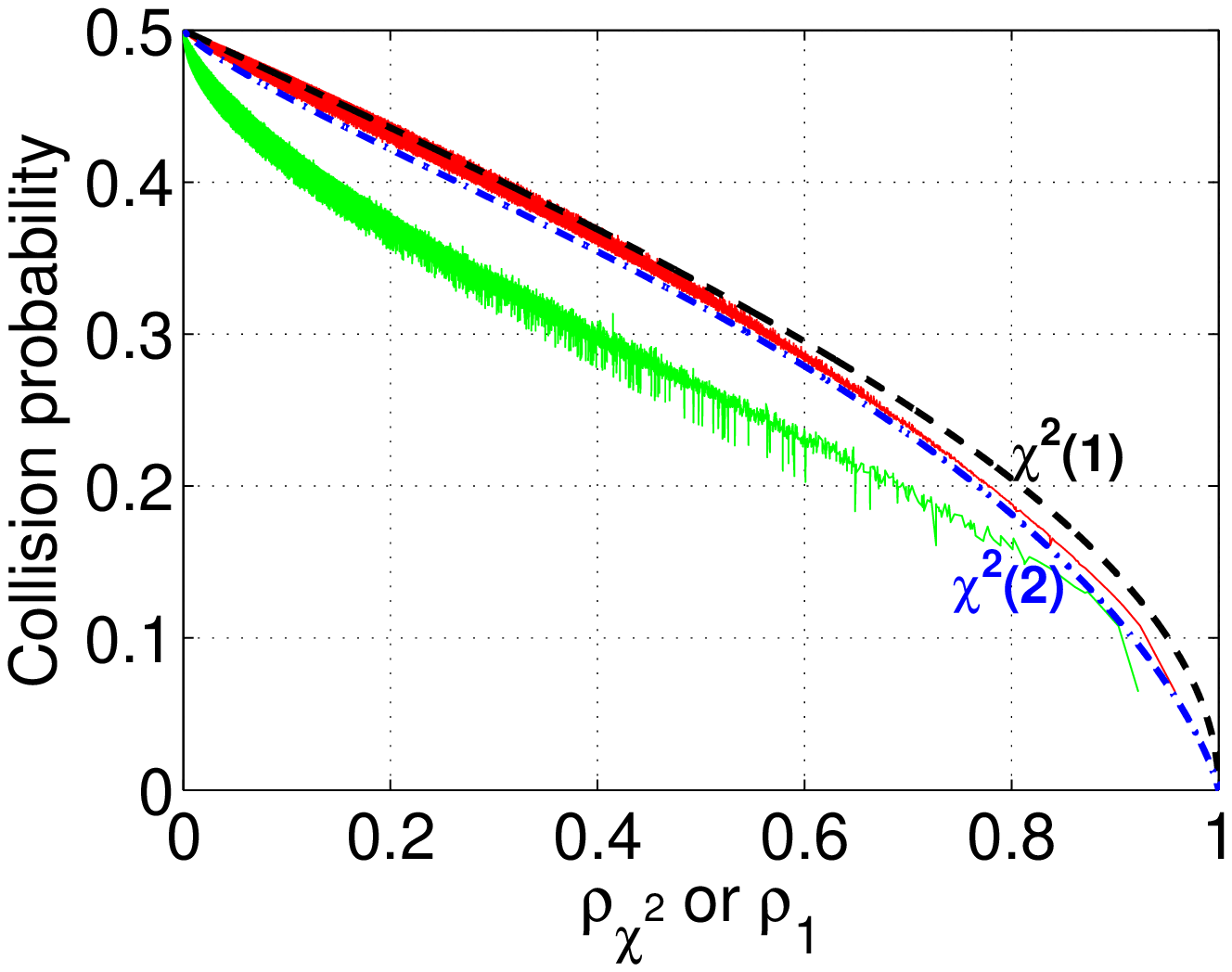}\hspace{0.2in}
\includegraphics[width=2.8in]{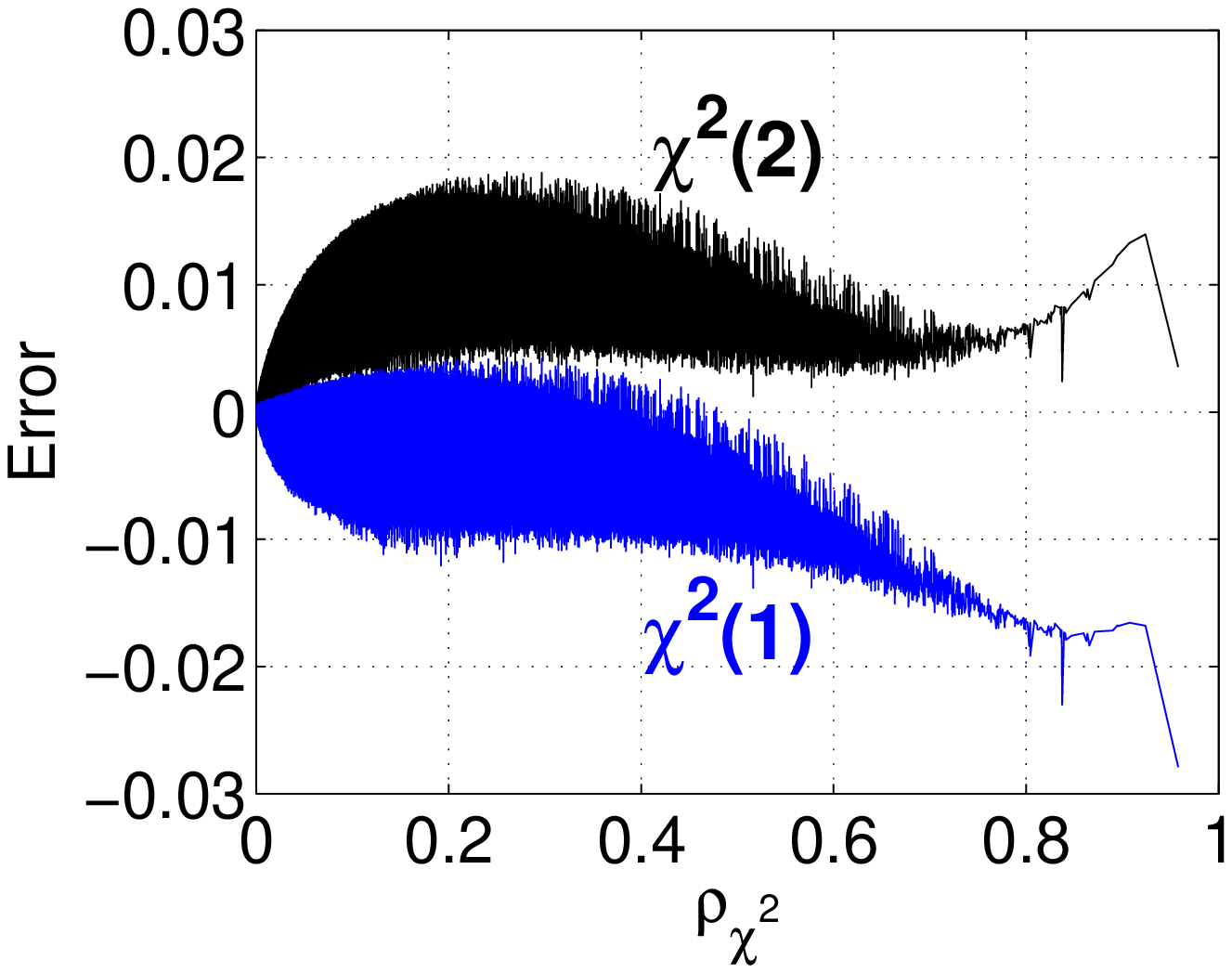}
\end{center}
\vspace{-0.2in}
\caption{We compute the empirical collision probabilities for all 3.6 million English word pairs. In the \textbf{left panel}, we plot the empirical collision probabilities against $\rho_1$ (lower, green if color is available) and $\rho_{\chi^2}$ (higher, red). The curves  confirm that the bound $\frac{1}{\pi}\cos^{-1}\rho_1$ is not tight (and the curve is not smooth). We plot the two $\chi^2$ approximations as dashed curves which largely match the empirical collision probabilities plotted against $\rho_{\chi^2}$. This confirms that the $\chi^2$ approximations are good. For smaller $\rho_{\chi^2}$ values, the first approximation $P_{\chi_2(1)}$ is slightly more accurate. For larger $\rho_{\chi^2}$ values, the second approximation $P_{\chi^2(2)}$ is more accurate. In the \textbf{right panel}, we plot the errors for both $P_{\chi_2(1)}$ and $P_{\chi_2(2)}$. 
}\label{fig_WordProb}
\end{figure}

\newpage

\section{Sign Cauchy Random Projections for  Classification}\label{sec_classification}

Our method provides an effective strategy for classification. For each (high-dimensional) data vector, using $k$ sign Cauchy projections, we   encode a negative sign as ``01'' and a positive  as ``10'' (i.e., a vector of length 2) and  concatenate   $k$ short vectors to form a new feature vector of length $2k$ (with exactly $k$ 1's). We  then feed the new dataset into a linear classifier (e.g., LIBLINEAR). Interestingly, this linear classifier  approximates a nonlinear kernel classifier based on the acos-$\chi^2$-kernel: $K(u,v) = 1-\frac{1}{\pi}\cos^{-1}\rho_{\chi^2}$. See  Figure~\ref{fig_AcosChi2SVM} for  the experiments on the same two datasets in Figure~\ref{fig_base_chi2}: {\em UCI-PEMS} and {\em MNIST-Small}.

\begin{figure}[h!]
\begin{center}
\includegraphics[width=2.8in]{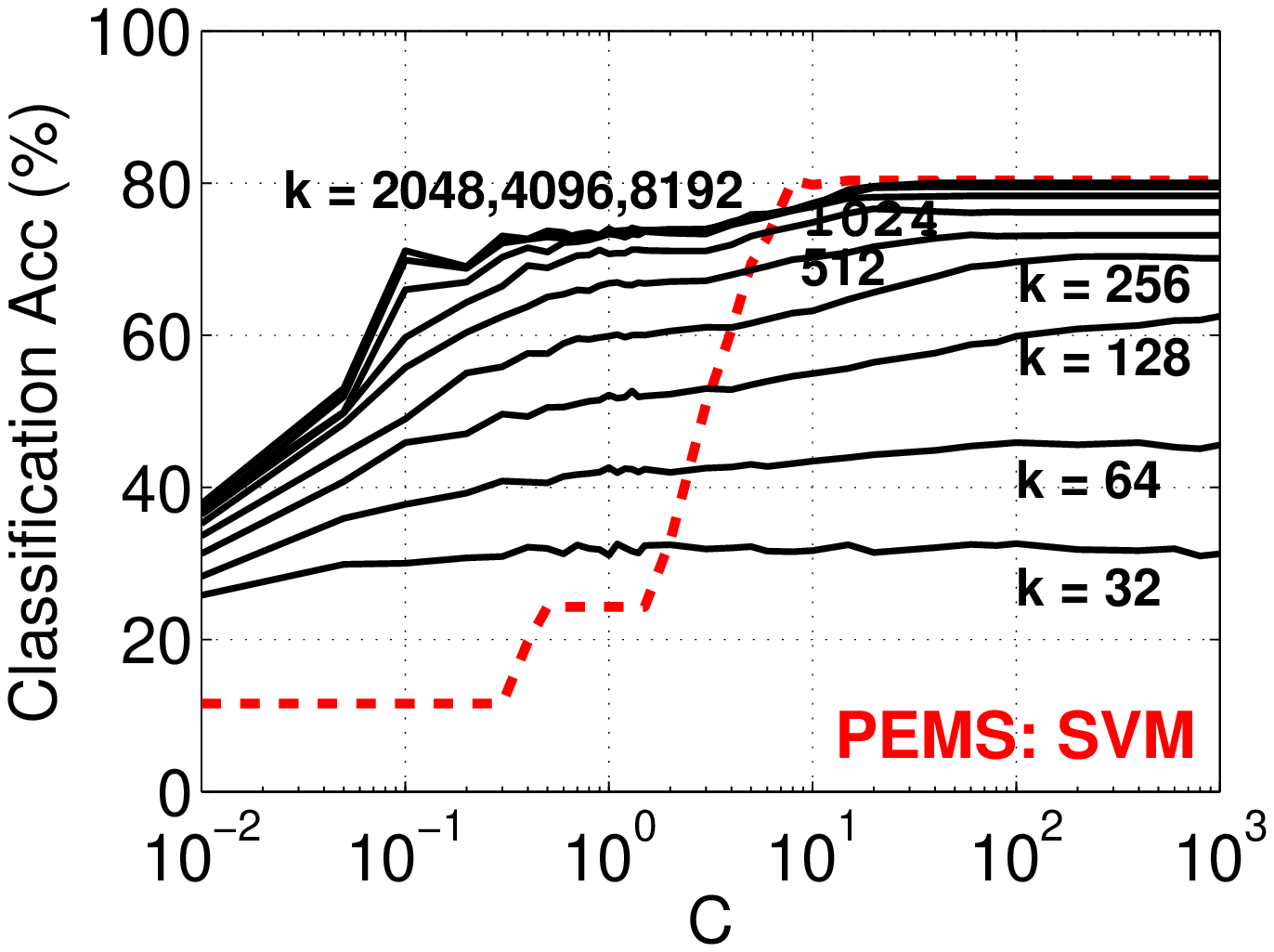}\hspace{0.3in}
\includegraphics[width=2.8in]{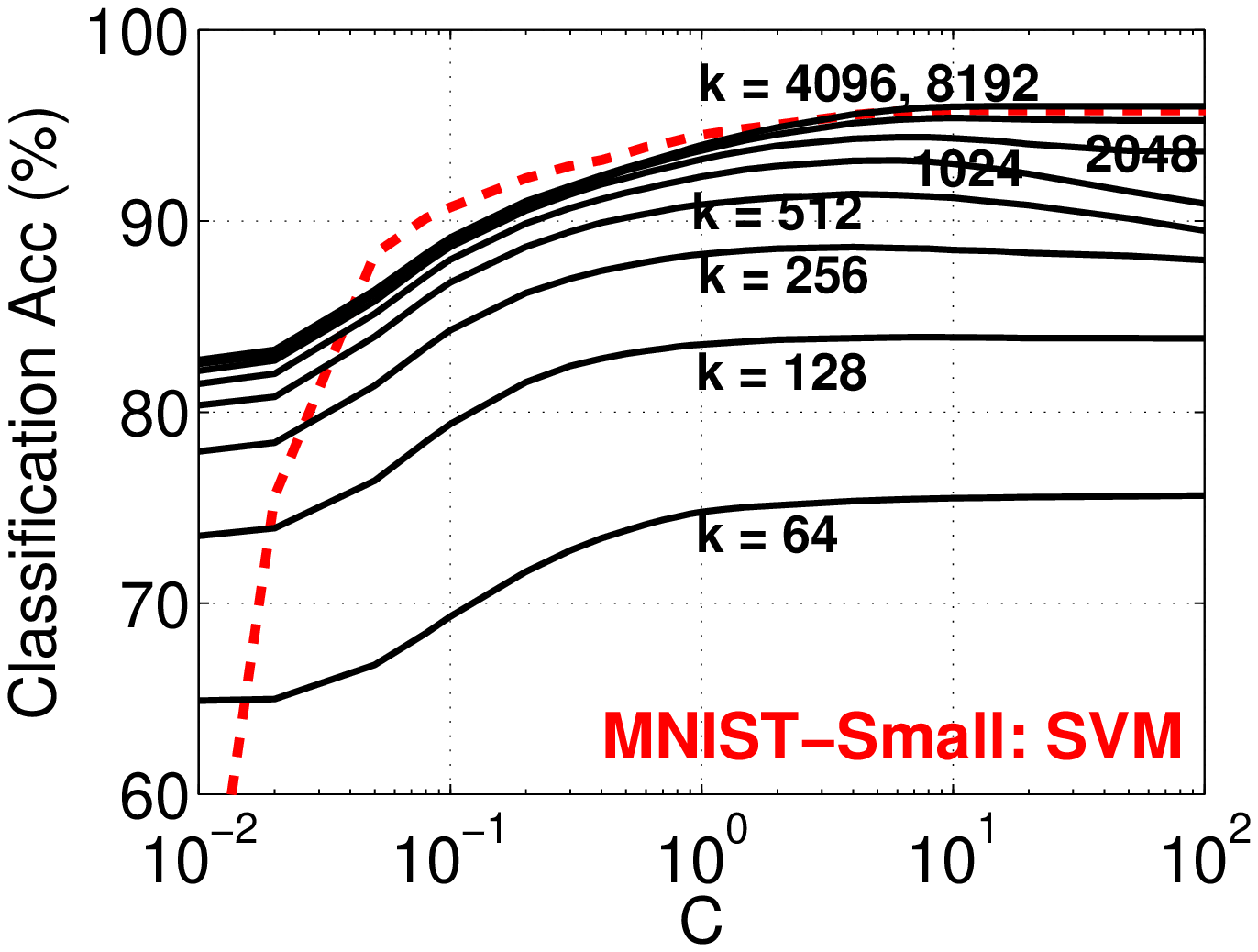}
\end{center}
\vspace{-0.2in}
\caption{The two dashed (red if color is available) curves are the classification results obtained using ``acos-$\chi^2$-kernel'' via the ``precomputed kernel'' functionality in LIBSVM. The solid (black) curves are the  accuracies using $k$ sign Cauchy projections and LIBLINEAR. The results confirm that the linear kernel from sign Cauchy projections can  approximate the nonlinear  acos-$\chi^2$-kernel.}\label{fig_AcosChi2SVM}
\end{figure}

Figure~\ref{fig_base_chi2} has already shown that, for the {\em UCI-PEMS} dataset, the $\chi^2$-kernel ($\rho_{\chi^2}$) can produce noticeably better classification results than the acos-$\chi^2$-kernel ($1-\frac{1}{\pi}\cos^{-1}\rho_{\chi^2}$). Although our method  does not directly approximate $\rho_{\chi^2}$, we can still estimate $\rho_{\chi^2}$ by assuming the collision probability is exactly $\mathbf{Pr}\left(\text{sign}(x)\neq \text{sign}(y)\right) = \frac{1}{\pi}\cos^{-1}\rho_{\chi^2}$ and then we can feed the estimated $\rho_{\chi^2}$ values into LIBSVM ``precomputed kernel'' for classification. Figure~\ref{fig_Chi2SVM} verifies that this method can also approximate the $\chi^2$ kernel with enough projections.

\begin{figure}[h!]
\begin{center}
\includegraphics[width=2.8in]{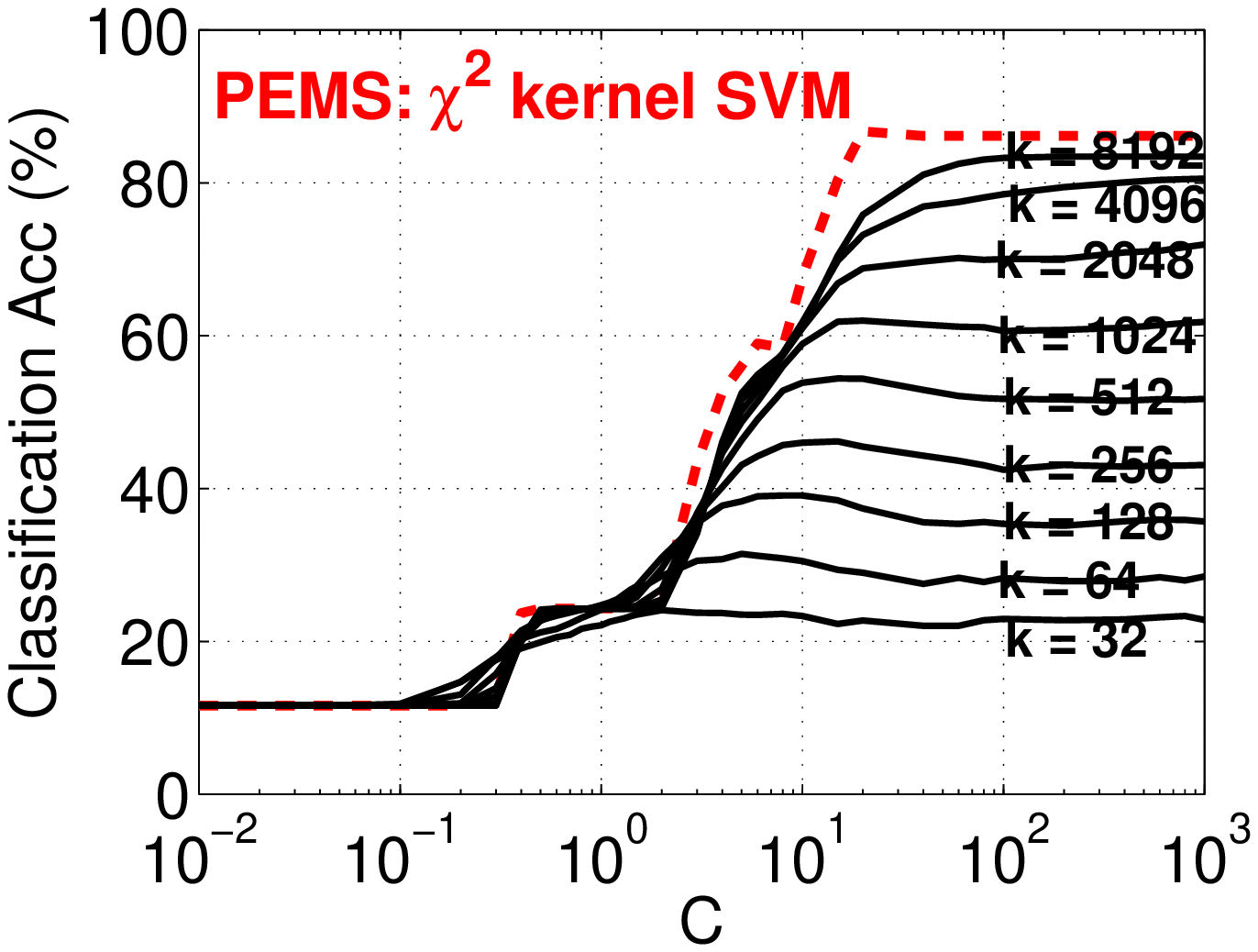}\hspace{0.3in}
\includegraphics[width=2.8in]{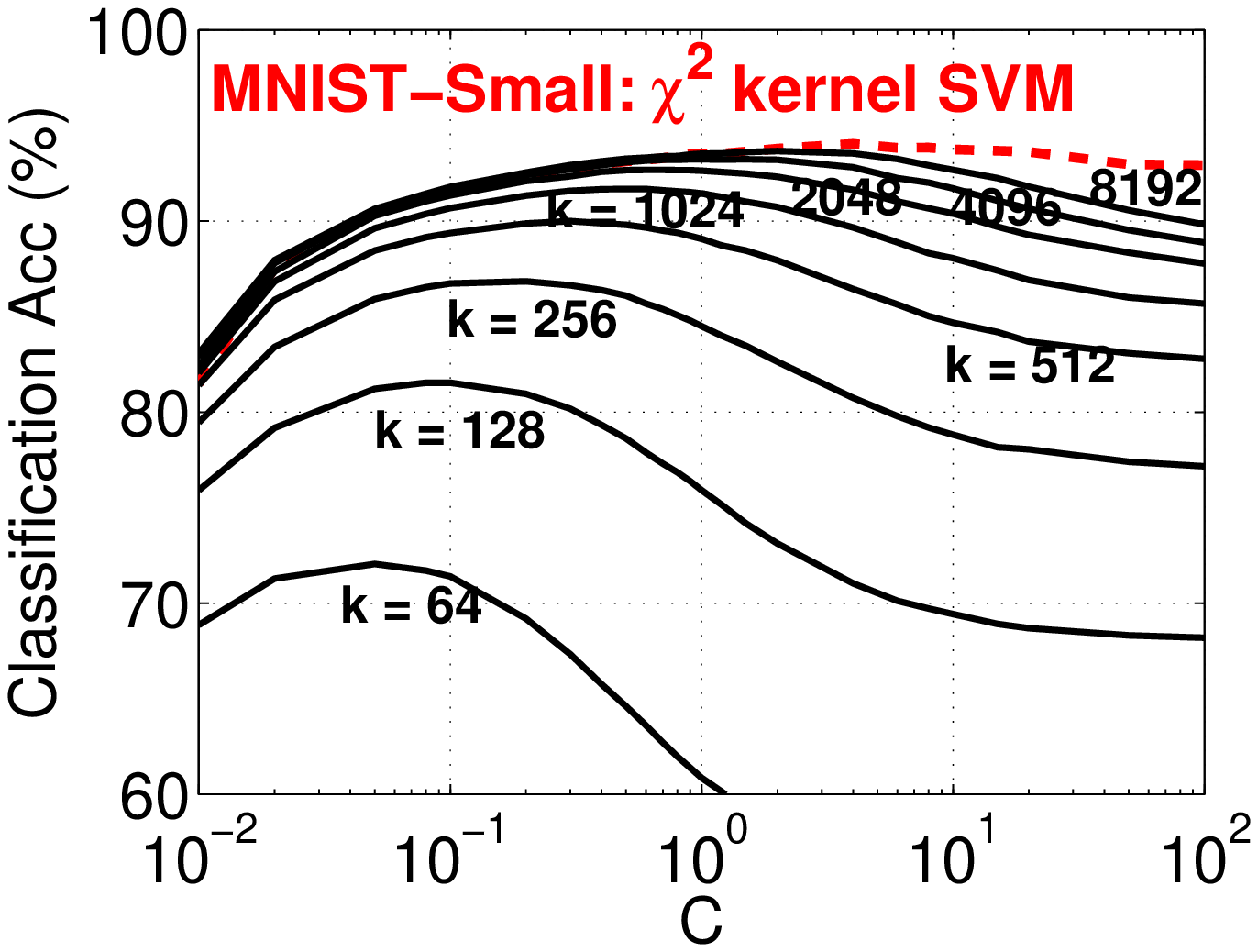}
\end{center}
\vspace{-0.2in}
\caption{\textbf{Nonlinear kernels}. The dashed curves are the classification results obtained using $\chi^2$-kernel and LIBSVM ``precomputed kernel'' functionality. We apply $k$ sign Cauchy projections and estimate $\rho_{\chi^2}$ assuming the collision probability is exactly $\frac{1}{\pi}\cos^{-1}\rho_{\chi^2}$ and then feed the estimated $\rho_{\chi^2}$ into LIBSVM again using the ``precomputed kernel'' functionality.}\label{fig_Chi2SVM}
\end{figure}

\section{Conclusion}\label{sec_conclusion}

The use of $\chi^2$ similarity is widespread in many machine learning applications, especially when features are generated from histograms, as common in natural language processing and computer vision. In fact, many prior studies~\cite{Article:Chapelle_99,Proc:Hein_AISTATS05,Proc:Jiang_CIVR07,Proc:Alexe_CVPR10,Proc:Wang_CVPR09,Proc:Vempati_BMVC10,Article:Vedaldi_PAMI12} have shown the advantage of using $\chi^2$ similarity compared to other measures such as $l_2$ or $l_1$ distances. Therefore, efficient algorithms for computing $\chi^2$ similarities will be of great interest in machine learning, computer vision, and information retrieval.

For large-scale applications with ultra-high-dimensional datasets, using the $\chi^2$ similarity becomes challenging for various practical reasons. Simply storing (and maneuvering) all the high-dimensional features would be difficult if there are a large number of observations (data points). Computing all pairwise $\chi^2$ similarities can be time-consuming and in fact we usually can not  materialize an all-pairwise similarity matrix if there are (e.g.,) merely $10^6$ data points. Furthermore, the $\chi^2$ similarity is nonlinear, making it difficult to take advantage of modern linear algorithms which are known to be very efficient, for example,~\cite{Proc:Joachims_KDD06,Proc:Shalev-Shwartz_ICML07,Article:Fan_JMLR08,URL:Bottou_SGD}. When  data are generated in a streaming fashion, computing $\chi^2$ similarities without storing the original data will be even  more challenging.\\

The method of {\em stable random projections}~\cite{Article:Indyk_JACM06,Proc:Li_SODA08} is popular for efficiently computing the $l_\alpha$ distances in massive (and possibly streaming) data. We propose {\em sign stable random projections} by storing only the signs (i.e., 1-bit) of the projected data  instead of real-valued numbers. The saving in storage would be a significant advantage. Also, the bits offer the indexing capability which allows efficient search. For example, we can build hash tables using the bits to achieve sublinear time near neighbor search, although this paper does not focus on near neighbor search. We can also build efficient linear classifiers using these bits, for  large-scale high-dimensional applications with {\em Big Data}.

To analyze sign stable random projections, a crucial task is to study the probability of collision (i.e., when the two signs differ). For general $0<\alpha\leq2$, we derive a theoretical  bound of the collision probability which is exact when $\alpha=2$. The bound is fairly sharp for $\alpha$ close to 2 (e.g., $\alpha\geq 1.5$). For $\alpha=1$ (i.e., Cauchy random projections), we find the $\chi^2$ approximation is  significantly more accurate. In addition, for binary data, we can analytically show that the errors from using the $\chi^2$ approximation are less than 0.0192. Experiments on real and simulated data confirm that our proposed $\chi^2$ approximations are very accurate.\\

We are enthusiastic about the practicality of {\em sign stable projections} in large-scale learning and search applications. The previous idea of using the signs from {\em normal random projections}  has been widely adopted in practice, for approximating correlations. Given  the wide spread  use of $\chi^2$ similarity,  the practical advantage of $\chi^2$ similarity  in text and vision applications, and the simplicity of our method, we expect the proposed  method will  be adopted by practitioners.\\

Finally, to conclude this paper, we should mention that the processing cost of conducting stable random projections can be dramatically reduced by {\em very sparse stable random projections}~\cite{Proc:Li_KDD07}. This will make our proposed method even more practical.

\appendix
\section{Proof of Theorem~\ref{thm_upper_bound}}\label{app_thm_upper_bound}

Apply stable random projections: $x = \sum_{i=1}^D u_i r_i$, $y=\sum_{i=1}^D v_i r_i$, $r_i \sim S(\alpha,0,1)$ i.i.d. Using the result in Lemma~\ref{lem_factorization}, we can factor $r_i$ into a product of two independent random variables:
\begin{align}\notag
r_i =T_i^{1/2}\times N_i, \ \ \ T_i \sim S\left(\alpha/2,1,\cos\left(\frac{\pi\alpha}{4}\right)\right), \ \ N_i \sim S(2,0,1)
\end{align}
Conditioning on $T_1, T_2, ..., T_D$, the collision probability becomes
\begin{align}\notag
\mathbf{Pr}\left(\text{sign}(x) \neq \text{sign}(y)|T_i, i=1, 2, ..., D\right) = \frac{1}{\pi}\cos^{-1}\left(\frac{\sum_{i=1}^D u_iv_iT_i}{\sqrt{\sum_{i=1}^Du_i^2T_i\sum_{i=1}^D v_i^2T_i}}\right)
\end{align}
and hence the unconditional probability can be written as
\begin{align}\notag
\mathbf{Pr}\left(\text{sign}(x) \neq \text{sign}(y)\right) = E\left(\frac{1}{\pi}\cos^{-1}\left(\frac{\sum_{i=1}^D u_iv_iT_i}{\sqrt{\sum_{i=1}^Du_i^2T_i\sum_{i=1}^D v_i^2T_i}}\right)\right)
\end{align}
When the data are nonnegative, i.e., $u_i\geq0, v_i\geq 0$, we have
\begin{align}\notag
&\sum_{i=1}^D u_iv_iT_i = \left(\sum_{i=1}^D u_i^{\alpha/2}v_i^{\alpha/2}\cos\left(\frac{\pi\alpha}{4}\right)\right)^{2/\alpha}\times S_1 \\\notag
&\sum_{i=1}^D u_i^2T_i =\left(\sum_{i=1}^D u_i^{\alpha}\cos\left(\frac{\pi\alpha}{4}\right)\right)^{2/\alpha}\times S_2, \\\notag
&\sum_{i=1}^D v_i^2T_i = \left(\sum_{i=1}^D u_i^{\alpha}\cos\left(\frac{\pi\alpha}{4}\right)\right)^{2/\alpha}\times S_3
\end{align}
where the random variables $S_1, S_2, S_3 \sim S(\alpha/2,1,1)$ are not independent.

As $\cos^{-1}(z)$ is a concave, monotonically decreasing function of $z\in[0,\ 1]$, we can bound the probability by Jensen's inequality as follows:
\begin{align}\notag
&\mathbf{Pr}\left(\text{sign}(x) \neq \text{sign}(y)\right)\\\notag
 =& E\left(\frac{1}{\pi}\cos^{-1}\left(\frac{\sum_{i=1}^D u_iv_iT_i}{\sqrt{\sum_{i=1}^Du_i^2T_i\sum_{i=1}^D v_i^2T_i}}\right)\right)\\\notag
\leq& \frac{1}{\pi}\cos^{-1}\left\{E\left(\frac{\sum_{i=1}^D u_iv_iT_i}{\sqrt{\sum_{i=1}^Du_i^2T_i\sum_{i=1}^D v_i^2T_i}}\right)\right\}\\\notag
=&\frac{1}{\pi}\cos^{-1}\left\{E\exp\left(\log\left(\frac{\sum_{i=1}^D u_iv_iT_i}{\sqrt{\sum_{i=1}^Du_i^2T_i\sum_{i=1}^D v_i^2T_i}}\right)\right)\right\}\\\notag
\leq&\frac{1}{\pi}\cos^{-1}\left\{\exp\left(E\log\left(\frac{\sum_{i=1}^D u_iv_iT_i}{\sqrt{\sum_{i=1}^Du_i^2T_i\sum_{i=1}^D v_i^2T_i}}\right)\right)\right\}\\\notag
=&\frac{1}{\pi}\cos^{-1}\left\{\exp\left(\frac{2}{\alpha}\log\left(\sum_{i=1}^D u_i^{\alpha/2}v_i^{\alpha/2}\right)- \frac{1}{\alpha}\log\left(\sum_{i=1}^D u_i^{\alpha}\right)-\frac{1}{\alpha}\log\left(\sum_{i=1}^D v_i^{\alpha}\right)+E\log \frac{S_1}{\sqrt{S_2S_3}} \right)\right\}\\\notag
= &\frac{1}{\pi}\cos^{-1}\left\{\left(\frac{\sum_{i=1}^D u_i^{\alpha/2}v_i^{\alpha/2}}{\sqrt{\sum_{i=1}^Du_i^\alpha\sum_{i=1}^D v_i^\alpha}}\right)^{2/\alpha}\right\}
=\frac{1}{\pi}\cos^{-1}\rho_\alpha
\end{align}
Note that $E\log \frac{S_1}{\sqrt{S_2S_3}} = E\log S_1 - 1/2E\log S_2 - 1/2E\log S_3 =0$ because $S_1, S_2, S_3$ follow the same distribution (although they are not independent). This completes the proof.

\section{Proof of Theorem~\ref{thm_binary_prob}}\label{app_thm_binary_prob}

For convenience, we define $a = |I_a|, \ \ b = |I_b|, \ \ c = |I_{c}|$, where
\begin{align}\notag
I_a = \{i|u_i>0, v_i=0\},\hspace{0.2in} I_b = \{i|v_i>0, u_i=0\},\hspace{0.2in} I_{c} = \{i|u_i>0, v_i>0\},
\end{align}

Assume binary data (before normalization), i.e.,
\begin{align}\notag
u_i = \frac{1}{|I_a|+|I_{c}|}=\frac{1}{a+c}, \ \ \forall i\in I_a\cup I_c,\hspace{0.5in}
v_i = \frac{1}{|I_b|+|I_{c}|}=\frac{1}{b+c}, \ \ \forall i\in I_b\cup I_c
\end{align}
The chi-square similarity $\rho_{\chi^2}$  becomes
$\rho_{\chi^2} = \sum_{i=1}^D \frac{2u_iv_i}{u_i+v_i} =\frac{2c}{a+b+2c}$ and hence
$\frac{\rho_{\chi^2}}{2-2\rho_{\chi^2}} = \frac{c}{a+b}$.

We apply Cauchy projections, i.e., $r_i, i=1$ to $D$ are i.i.d. standard Cauchy.
\begin{align}\notag
&x = \sum_{i=1}^D u_i r_i = \frac{1}{a+c}\sum_{i\in I_a}r_i +\frac{1}{a+c}\sum_{i\in I_c}r_i = \frac{a}{a+c}R_1 + \frac{c}{a+c}R_3 \\\notag
&y = \sum_{i=1}^D v_i r_i = \frac{1}{b+c}\sum_{i\in I_b}r_i +\frac{1}{b+c}\sum_{i\in I_c}r_i  = \frac{b}{b+c}R_2 + \frac{c}{b+c}R_3
\end{align}
where $R_1, R_2, R_3$ are i.i.d. standard Cauchy variables.
\begin{align}\notag
&\mathbf{Pr}\left(x <0, y<0\right)\\\notag
=& \mathbf{Pr}\left(\frac{a}{a+c}R_1 + \frac{c}{a+c}R_3 <0, \frac{b}{b+c}R_2 + \frac{c}{b+c}R_3<0\right)\\\notag
=&\mathbf{Pr}\left(R_1 <-\frac{c}{a}R_3, R_2 < -\frac{c}{b}R_3\right)\\\notag
=&\frac{1}{\pi^2}E\left\{\left(\tan^{-1}\left(-\frac{c}{a}R_3\right)+\frac{\pi}{2} \right)
\left(\tan^{-1}\left(-\frac{c}{b}R_3\right)+\frac{\pi}{2} \right)\right\}\\\notag
\end{align}
By symmetry
\begin{align}\notag
&\mathbf{Pr}\left(x >0, y>0\right) =\mathbf{Pr}\left(x<0, y<0\right)  
\end{align}
and
\begin{align}\notag
\mathbf{Pr}\left(\text{sign}(x)\neq \text{sign}(y)\right) =&1-\frac{2}{\pi^2}E\left\{\left(\tan^{-1}\left(-\frac{c}{a}R_3\right)+\frac{\pi}{2} \right)
\left(\tan^{-1}\left(-\frac{c}{b}R_3\right)+\frac{\pi}{2} \right)\right\}\\\notag
=&\frac{1}{2}-\frac{2}{\pi^2}E\left\{\tan^{-1}\left(\frac{c}{a}|R_3|\right)\tan^{-1}\left(\frac{c}{b}|R_3|\right)\right\}
\end{align}


\section{Proof of Lemma~\ref{lem_ErrBinary}}\label{app_lem_ErrBinary}

The task is to prove the property of $Z(t_1,t_2)$, where $t_1\geq 0$, $t_2\geq 0$, and
\begin{align}\notag
Z(t_1,t_2) =&-\frac{2}{\pi^2}E\left\{\tan^{-1}\left(\frac{1}{t_1}|R|\right)\tan^{-1}\left(\frac{1}{t_2}|R| \right) \right\}
+\frac{1}{\pi} E\left\{\tan^{-1}\left(\frac{1}{t_1+t_2}|R|\right)\right\}
\end{align}
Here $R$ is a standard Cauchy random variable with density $f_R(r) = \frac{1}{\pi}\frac{1}{1+r^2}$.

At the boundaries (i.e., $t_1=0$, or $t_2=0$, or $t_1=\infty$, or $t_2=\infty$), we have $Z(t_1,t_2)=0$. We hope to find other extremum points by checking the first derivatives:
\begin{align}\notag
&\frac{\partial Z(t_1,t_2)}{\partial t_1} = \frac{2}{\pi^2}E\left\{\frac{|R|}{t_1^2+R^2}
\tan^{-1}\left(\frac{1}{t_2}|R|\right)
\right\} - \frac{1}{\pi} E\left\{\frac{|R|}{(t_1+t_2)^2+R^2}\right\}\\\notag
&\frac{\partial Z(t_1,t_2)}{\partial t_2} = \frac{2}{\pi^2}E\left\{\frac{|R|}{t_2^2+R^2}
\tan^{-1}\left(\frac{1}{t_1}|R|\right)
\right\} - \frac{1}{\pi} E\left\{\frac{|R|}{(t_1+t_2)^2+R^2}\right\}
\end{align}
The integrals can be analytically calculated.
\begin{align}\notag
&E\left\{\frac{|R|}{(t_1+t_2)^2+R^2}\right\}=\frac{1}{\pi}\int_0^\infty \frac{2r}{(t_1+t_2)^2+r^2}\frac{1}{1+r^2} dr
=\frac{1}{\pi}\int_0^\infty \frac{1}{(t_1+t_2)^2+r}\frac{1}{1+r} dr\\\notag
=&\frac{1}{\pi}\left.\frac{1}{(t_1+t_2)^2-1}\log\frac{t+1}{t+(t_1+t_2)^2}\right|_0^\infty = \frac{1}{\pi}\frac{\log((t_1+t_2)^2)}{(t_1+t_2)^2-1}
\end{align}
Using the integral table~\cite[4.535.14]{Book:Gradshteyn_07}, we obtain
\begin{align}\notag
&E\left\{\frac{|R|}{t_1^2+R^2}\tan^{-1}\left(\frac{1}{t_2}|R|\right)\right\} =
\frac{1}{\pi}\int_{-\infty}^{\infty} \tan^{-1}\left(\frac{r}{t_2}\right)\frac{r}{r^2+t_1^2}\frac{1}{1+r^2}dr
=  \frac{1}{t_1^2-1}\log\frac{t_1+t_2}{1+t_2}
\end{align}

Therefore,
\begin{align}\notag
& \frac{\partial Z(t_1,t_2)}{\partial t_1}= \frac{2}{\pi^2}\frac{1}{t_1^2-1}\log\frac{t_1+t_2}{1+t_2} - \frac{2}{\pi^2}\frac{\log(t_1+t_2)}{(t_1+t_2)^2-1}\\\notag
&\frac{\partial Z(t_1,t_2)}{\partial t_2} = \frac{2}{\pi^2}\frac{1}{t_2^2-1}\log\frac{t_1+t_2}{1+t_1} - \frac{2}{\pi^2}\frac{\log(t_1+t_2)}{(t_1+t_2)^2-1}
\end{align}
Setting $\frac{\partial Z(t_1,t_2)}{\partial t_1}=0$ and $\frac{\partial Z(t_1,t_2)}{\partial t_2}=0$ yields
\begin{align}\notag
& \frac{1}{t_1^2-1}\log\frac{t_1+t_2}{1+t_2} - \frac{\log(t_1+t_2)}{(t_1+t_2)^2-1}=0\\\notag
&\frac{1}{t_2^2-1}\log\frac{t_1+t_2}{1+t_1} - \frac{\log(t_1+t_2)}{(t_1+t_2)^2-1}=0
\end{align}
which means we must have
\begin{align}\notag
g(t_1,t_2) = (t_1^2-1) \log\frac{t_1+t_2}{1+t_1} - (t_2^2-1) \log\frac{t_1+t_2}{1+t_2} =0
\end{align}
which obviously holds when $t_1=t_2$. We just need to make sure   there are no other solutions with $t_1\neq t_2$.

Note that, for fixed $t_2<1$, $g(t_1,t_2)$ is a concave function in $t_1$; and for fixed $t_2>1$, $g(t_1,t_2)$ is a convex function in $t_1$. To see this fact, we compute the derivatives of $g(t_1,t_2)$:
 \begin{align}\notag
&\frac{\partial g(t_1,t_2)}{\partial t_1} = 2t_1 \log\frac{t_1+t_2}{1+t_1} + (t_1^2-1)\left(\frac{1}{t_1+t_2} - \frac{1}{1+t_1}\right) - \frac{t_2^2-1}{t_1+t_2}= 2t_1 \log\frac{t_1+t_2}{1+t_1} -t_2+1\\\notag
&\frac{\partial^2 g(t_1,t_2)}{\partial t_1^2} = 2 \log\frac{t_1+t_2}{1+t_1} + 2t_1 \left(\frac{1}{t_1+t_2} - \frac{1}{1+t_1}\right)
\end{align}
For fixed $t_2$, we have $\left.\frac{\partial^2 g(t_1,t_2)}{\partial t_1^2}\right|_{t_1=\infty} = 0$. We can check the third derivative:
\begin{align}\notag
&\frac{\partial^3 g(t_1,t_2)}{\partial t_1^3} = 4\left(\frac{1}{t_1+t_2} - \frac{1}{1+t_1}\right) + 2t_1 \left(-\frac{1}{(t_1+t_2)^2} + \frac{1}{(1+t_1)^2}\right) = 2\frac{(1-t_2)(t_1+2t_2+t_1t_2)}{(t_1+t_2)^2(1+t_1)^2}
\end{align}
which means that, if $t_2<1$, then $\frac{\partial^3 g(t_1,t_2)}{\partial t_1^3}\geq 0$ (i.e., $\frac{\partial^2 g(t_1,t_2)}{\partial t_1^2}\leq 0$), and if  $t_2>1$ then  $\frac{\partial^3 g(t_1,t_2)}{\partial t_1^3}\leq 0$ (i.e., $\frac{\partial^2 g(t_1,t_2)}{\partial t_1^2}\geq 0$).\\

For fixed $t_2$, the function $g(t_1,t_2)$ is either convex or concave, which means $g(t_1,t_2)=0$ has at most two solutions. In the meanwhile, we know that if $t_1=t_2$, or $t_1=1$, then  $g(t_1,t_2)=0$.

Now it remains to check the case of $t_1=1$. When $t_1\rightarrow1$, we have
\begin{align}\notag
& \frac{\partial Z(t_1,t_2)}{\partial t_1}\rightarrow \frac{2}{\pi^2}\frac{1}{(t_1+1)(t_2+1)}- \frac{2}{\pi^2}\frac{\log(1+t_2)}{(1+t_2)^2-1}\neq 0
\end{align}

This way, we can conclude that the extremum points can only happen at $t_1=t_2$ and hence we only need to study a one-dimensional function.
\begin{align}\label{eqn_h(t)}
&h(t) = \frac{\pi^2}{2}\left.\frac{\partial Z(t_1,t_2)}{\partial t_1}\right|_{t_1=t_2=t}= \frac{1}{t^2-1}\log\frac{2t}{1+t} - \frac{\log(2t)}{(2t)^2-1}
\end{align}
$h(t)$ is plotted in Figure~\ref{fig_h(t)}. The solution to $h(t) = 0$ is $t^*=2.7793457703$. Moreover, $h(t)>0$ if $t<t^*$ and $h(t)<0$ if $t>t^*$. This means the maximum of $Z(t_1,t_2)$ is attained at $t_1=t_2=t^*$.
\begin{figure}[h!]
\begin{center}
\includegraphics[width=2.5in]{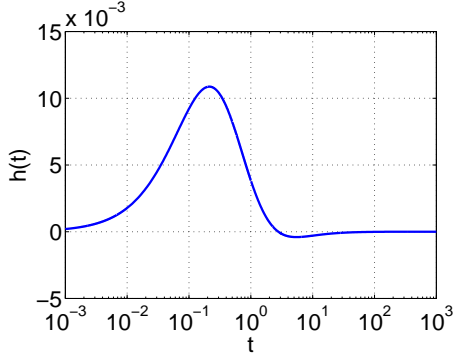}
\end{center}
\vspace{-0.2in}
\caption{The function $h(t)$ as defined in (\ref{eqn_h(t)}). It is positive when $t<t^*=2.7793457703$ and negative when $t>t^*$.}\label{fig_h(t)}

\end{figure}

Now it remains to compute the maximum value of $Z(t)$, where
\begin{align}\notag
Z(t) =& Z(t_1=t,t_2=t)\\\notag
 =& -\frac{2}{\pi^2}E\left\{\tan^{-1}\left(\frac{1}{t}|R|\right)\tan^{-1}\left(\frac{1}{t}|R| \right) \right\}
+\frac{1}{\pi} E\left\{\tan^{-1}\left(\frac{1}{2t}|R|\right)\right\}\\\notag
=&\int_0^{\infty} \left\{-\frac{2}{\pi^2}\left(\tan^{-1}\left(\frac{r}{t}\right)\right)^2 +\frac{1}{\pi}\tan^{-1}\left(\frac{r}{2t}\right)\right\} \frac{2}{\pi}\frac{1}{1+r^2}dr
\end{align}
which reaches the maximum at $t =t^*$.

{

}

\end{document}